\documentclass[12pt, final]{l4dc2022}

\title[Physics-Constrained Neural ODEs for Dynamical Systems Modeling]{Neural Networks with Physics-Informed \\ Architectures and Constraints for Dynamical Systems Modeling}

\usepackage{times}
\usepackage[utf8]{inputenc}

\usepackage{algorithm}
\usepackage{algorithmic}

\usepackage{pgfplots}
\DeclareUnicodeCharacter{2212}{−}
\usepgfplotslibrary{groupplots,dateplot}
\usetikzlibrary{patterns,shapes.arrows}
\pgfplotsset{compat=newest}
\usepackage{hyperref}

\usepackage{adjustbox}
\usepackage{multirow}
\usepackage{pbox}

\usepackage{tikz}
\usetikzlibrary{shapes, arrows.meta, positioning}

\usepgfplotslibrary{groupplots}
\usepgfplotslibrary{fillbetween}
\usetikzlibrary{arrows,decorations.pathmorphing,positioning,fit,trees,shapes,shadows,automata,calc} 
\usetikzlibrary{patterns,arrows,arrows.meta,calc,shapes,shadows,decorations.pathmorphing,decorations.pathreplacing,automata,shapes.multipart,positioning,shapes.geometric,fit,circuits,trees,shapes.gates.logic.US,fit, matrix,arrows.meta, quotes}
\usetikzlibrary{backgrounds,scopes}
\usepackage{ulem}

\usepackage{tabu}

\newcommand\freefootnote[1]{%
  \let\thefootnote\relax%
  \footnotetext{#1}%
  \let\thefootnote\svthefootnote%
}

\author{\Name{Franck Djeumou}\thanks{These authors contributed equally. \\ Project code is available at \href{https://github.com/wuwushrek/physics_constrained_nn}{https://github.com/wuwushrek/physics\_constrained\_nn}.}\textsuperscript{\rm 1} \Email{fdjeumou@utexas.edu}\\
    \Name{Cyrus Neary}\footnotemark[1]\textsuperscript{\rm 1} \Email{cneary@utexas.edu}\\
    \Name{Eric Goubault}\textsuperscript{\rm 2} \Email{goubault@lix.polytechnique.fr}\\
    \Name{Sylvie Putot}\textsuperscript{\rm 2} \Email{putot@lix.polytechnique.fr}\\
    \Name{Ufuk Topcu}\textsuperscript{\rm 1} \Email{utopcu@utexas.edu}\\
    \addr \textsuperscript{\rm 1} The University of Texas at Austin, United States\\
    \addr \textsuperscript{\rm 2} LIX, CNRS, \'Ecole Polytechnique, Institut Polytechnique de Paris, France
 }

\begin{document}

% Custom command
\newcommand{\fknown}{\mathrm{F}}
\newcommand{\funk}{\mathrm{g}}
\newcommand{\constr}{\Psi}

% System Dynamics
\newcommand{\state}{x}
\newcommand{\stateSpace}{\mathcal{X}}
\newcommand{\dynFun}{f}
\newcommand{\controlInput}{u}
\newcommand{\controlSpace}{\mathcal{U}}
\newcommand{\trajectory}{\tau}
\newcommand{\stateDim}{n}
\newcommand{\controlDim}{m}

\newcommand{\unknownTerm}{g}
\newcommand{\vectorField}{F}
\newcommand{\numUnknownTerms}{d}

% Neural Networks
\newcommand{\nnParamsAll}{\Theta}
\newcommand{\nnParams}{\theta}
\newcommand{\numParams}{k}

% Training
\newcommand{\predNextState}{\Gamma}
\newcommand{\loss}{\mathcal{J}}
\newcommand{\dataset}{\mathcal{D}}
\newcommand{\datasetDim}{|\dataset|}
\newcommand{\numTrainingTrajectories}{i}
\newcommand{\trajectoryTimeHorizon}{T}
\newcommand{\rollout}{n_r}
\newcommand{\odesolve}{\mathrm{ODESolve}}

% Constraints
\newcommand{\inequalConstr}{\Psi}
\newcommand{\equalConstr}{\Phi}
\newcommand{\constrDomain}{\mathcal{C}}
\newcommand{\numInequalConstr}{l}
\newcommand{\numEqualConstr}{v}
\newcommand{\collocationPoints}{\Omega}
\newcommand{\numCollocationPoints}{|\collocationPoints|}
\newcommand{\augmentedLagrangian}{\mathcal{L}}
\newcommand{\constrPenalty}{\mu}
\newcommand{\lagrangeVar}{\lambda}
\newcommand{\lagrangeVarAll}{\Lambda}
\newcommand{\totalNumEqualConstr}{N_{\equalConstr}}
\newcommand{\totalNumInequalConstr}{N_{\inequalConstr}}

\newcommand{\collectionUnknownTerms}{G_{\nnParamsAll}}

% algorithm
\newcommand{\constrTol}{\epsilon}

% dynamics equations
\newcommand{\massMat}{M}
\newcommand{\corForce}{C}
\newcommand{\actuationForce}{\tau}
\newcommand{\contactForce}{\mathcal{F}}
\newcommand{\jacobian}{J}

\newpage

\maketitle

\begin{abstract}
Effective inclusion of physics-based knowledge into deep neural network models of dynamical systems can greatly improve data efficiency and generalization.
Such a priori knowledge might arise from physical principles (e.g., conservation laws) or from the system's design (e.g., the Jacobian matrix of a robot), even if large portions of the system dynamics remain unknown.
We develop a framework to learn dynamics models from trajectory data while incorporating a priori system knowledge as inductive bias.
More specifically, the proposed framework uses physics-based side information to inform the structure of the neural network itself, and to place constraints on the values of the outputs and the internal states of the model.
It represents the system's vector field as a composition of known and unknown functions, the latter of which are parametrized by neural networks.
The physics-informed constraints are enforced via the augmented Lagrangian method during the model's training.
We experimentally demonstrate the benefits of the proposed approach on a variety of dynamical systems -- including a benchmark suite of robotics environments featuring large state spaces, non-linear dynamics, external forces, contact forces, and control inputs.
By exploiting a priori system knowledge during training, the proposed approach learns to predict the system dynamics two orders of magnitude more accurately than a baseline approach that does not include prior knowledge, given the same training dataset.
\end{abstract}


\section{Introduction}
\label{sec:introduction}

Owing to their tremendous capability to learn complex relationships from data, neural networks offer promise in their ability to model unknown dynamical systems from trajectory observations.
Such models of system dynamics can then be used to synthesize control strategies, to perform model-based reinforcement learning, or to predict the future values of quantities of interest.

However, purely data-driven approaches to learning can result in poor data efficiency and in model predictions that violate physical principles. 
These deficiencies become particularly emphasized when the training dataset is relatively small -- the neural network must learn to approximate a high-dimensional and non-linear map from a limited number of state trajectories. 
This limited reliability in the scarce data regime can render neural-network-based dynamics models impractical for the aforementioned applications.

On the other hand, useful a priori system knowledge is often available, even in circumstances when the exact dynamics remain unknown. 
Such knowledge might stem from a variety of sources -- basic principles of physics, geometry constraints arising from the system's design, or empirically validated invariant sets in the state space.
The central thesis of this paper is that effective inclusion of a priori knowledge into the training of deep neural network models of dynamical systems can greatly improve data efficiency and model generalization to previously unseen regions of the state space, while also ensuring that the learned model respects physical principles.
\begin{figure*}[t]
    \centering
    \input{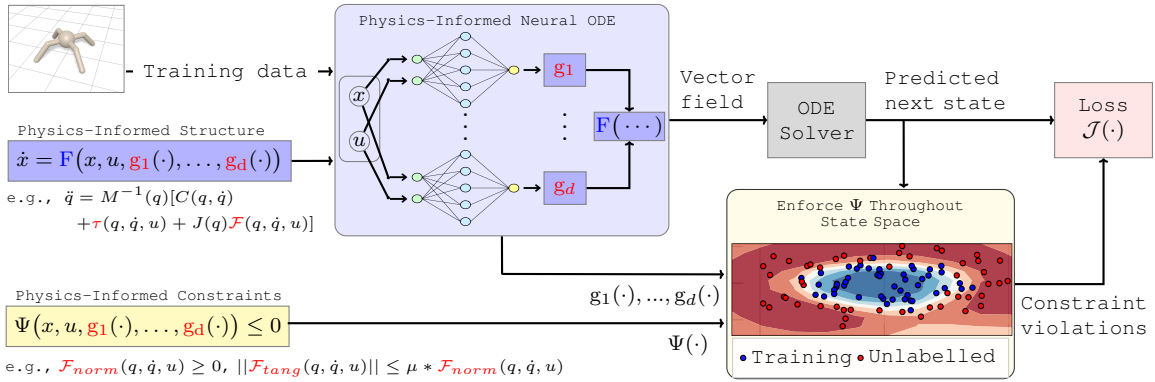}
    \vspace{-7mm}
    \caption{
    An illustration of the proposed framework.
    A priori physics knowledge is captured by a structured representation of the vector field (\textit{blue}) -- unknown components are represented by neural networks (\textit{red}). Physics-based constraints are enforced on the outputs of the model not only on the labeled training datapoints, but also on any unlabeled points within the state space where the constraints are known to hold true (\textit{yellow}).
    }
    \vspace{-5mm}
    \label{fig:approach_illustration}
\end{figure*} 

We develop a framework to incorporate a wide variety of potential sources of a priori knowledge into data-driven models of dynamical systems. 
The framework uses neural networks to parametrize the dynamical system's vector field and uses numerical integration schemes to predict future states, as opposed to parametrizing a model that predicts the next state directly.
Physics-based knowledge is incorporated as inductive bias in the model of the vector field via two distinct mechanisms:

\paragraph{1. Physics-informed model structure.}
We represent the system's vector field as compositions of unknown terms that are parametrized as neural networks and known terms that are derived from a priori knowledge.
For example, Figure \ref{fig:approach_illustration} illustrates a robotics environment in \textit{Brax} \citep{freeman2021brax} whose equations of motion we assume to be partially known; the robot's mass matrix is available, while the remaining terms in its equations of motion (i.e. its actuation and contact forces) must be learned.
Neural networks representing the unknown terms are composed with the known mass matrix to obtain a model of the system's vector field, as is illustrated in Figure \ref{fig:approach_illustration}.

\paragraph{2. Physics-informed constraints.}
We additionally enforce physics-based constraints on the values of the outputs and the internal states of the model.
Such constraints could, for example, encode known system equilibria, invariants, or symmetries in the dynamics.
We note that while only a limited number of datapoints may be available for supervised learning, constraints derived from a priori knowledge will hold over large sub-sets of the state space, and potentially over the state space in its entirety.
Ensuring that the learned model satisfies the relevant constraints throughout the state space, while also fitting the available trajectory data, leads to a semi-supervised learning scheme that generalizes the available training data to the unlabeled portions of the state space.

\vspace{2mm}

\begin{figure}[t]
    \centering
    \input{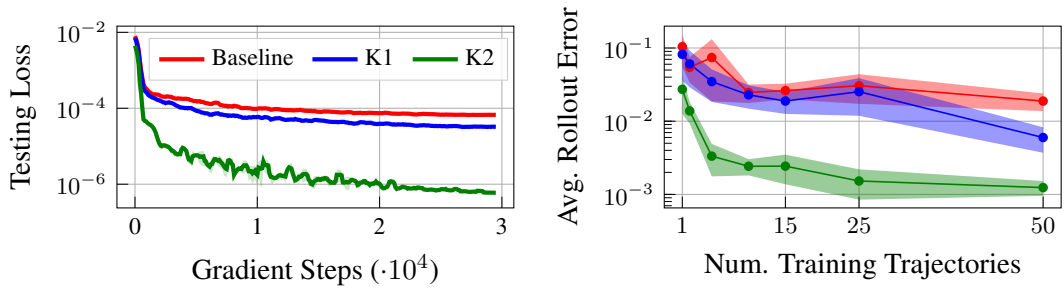}
    \vspace{-3mm}
    \caption{Incorporation of a priori knowledge results in a significant improvement in data efficiency and out-of-sample prediction accuracy.
    \textit{K1} corresponds to the incorporation of basic vector field knowledge: $\frac{dx}{dt} = v$.
    \textit{K2} corresponds to knowledge of the mass matrix of the robot.
    The \textit{Testing Loss} plot illustrates the model's prediction errors during training.
    The \textit{Avg. Rollout Error} plot illustrates a measure of the error over an entire predicted trajectory beginning from an out-of-sample initial state, after training has converged.}
    \vspace{-5mm}
    \label{fig:intro_plot}
\end{figure}

We train the model on time-series data of state and control input observations.
The physics-informed model of the vector field is integrated in time to obtain predictions of the future state.
These predictions are in turn compared against the true future state values obtained from training data to define the model's loss.
During the model's training, we enforce the physics-informed constraints on a collection of points in the state space, both labeled and unlabeled, via the augmented Lagrangian method \citep{lu2021physics, hestenes1969multiplier}. 

We experimentally demonstrate the effectiveness of the proposed approach on a double pendulum system as well as on a suite of controlled rigid multi-body systems simulated in Brax \citep{freeman2021brax}.
This suite includes systems with nonlinear dynamics, control inputs, contact forces, and states with hundreds of dimensions.
We are the first to consider such a suite of experiments in the context of physics-informed neural networks.
We consider symmetries in the system vector fields, knowledge of the mass and Jacobian matrices of the robotic systems, and constraints that encode the laws of friction to help learn contact forces.
We examine the extent to which these different sources of side information aid in improving the data efficiency and the out-of-sample prediction accuracy of the learned models.
By exploiting a priori system knowledge during training, the proposed approach learns to predict system dynamics more than two orders of magnitude more accurately than a baseline approach that does not include any prior knowledge, as illustrated in Figure \ref{fig:intro_plot}.
The experiments demonstrate the viability of the proposed framework for the efficient learning of complex systems that are often used as challenging benchmark environments for deep reinforcement learning and model-based control \citep{duan2016benchmarking, freeman2021brax}.
\section{Physics-Informed Neural Architectures}

We consider arbitrary non-linear dynamics that can be expressed in terms of an ordinary differential equation (ODE) of the form \(\dot{\state} = h(\state, \controlInput)\).
The state $\state : \mathbb{R}_+ \mapsto \stateSpace$ is a continuous-time signal, the control input  $\controlInput : \mathbb{R}_+ \mapsto \controlSpace$ is a (possibly non-continuous) signal of time, and the vector field $h : \stateSpace \times \controlSpace \mapsto \mathbb{R}^\stateDim$ is assumed to be unknown and possibly non-linear. 
Here, $\stateDim$ denotes the dimension of the state. 
We note that it is straightforward to explicitly include time dependency in the ODE representing the system dynamics, however, we omit it for clarity and notational simplicity. 

Throughout the paper, we assume that a finite dataset $\dataset$ of system trajectories -- time-series data of states and control inputs -- is available, in lieu of the system's model. 
That is, we are given a set $\dataset = \{ \trajectory_1, \hdots, \trajectory_{\datasetDim}\}$ of trajectories $\trajectory = \{(\state_0, \controlInput_0), (\state_1, \controlInput_1), \hdots,(\state_{\trajectoryTimeHorizon}, \controlInput_{\trajectoryTimeHorizon})\}$, where $\state_i = \state(t_i)$ is the state at time $t_i$, $\controlInput_i = \controlInput(t_i)$ is the control input applied at time $t_i$, and $t_0 < t_1 < \ldots < t_{\trajectoryTimeHorizon}$ is an increasing sequence of points in time.
We seek to learn a function to predict future state values $\state_{k+1},\hdots,\state_{k+\rollout+1}$, given the current state value $\state_k$ and a sequence of control inputs $\controlInput_k,\hdots,\controlInput_{k+\rollout}$.

We propose to parametrize and to learn the unknown vector field $h$, as opposed to learning a prediction function that maps directly from a history of states and control inputs to the predicted next state. 
This choice to parametrize a model of the vector field is primarily motivated by the two following points: 
(1) a priori knowledge derived from physical principles is typically most easily expressed in terms of the system's vector field;
(2) \cite{chen2018neural} recently demonstrated that parametrizing the vector field results in models that are able to approximate time-series data with more accuracy than approaches that directly estimate the next state.
We thus build a framework capable of learning the vector field $h$ from the dataset $\dataset$ of trajectories, while also taking advantage of prior knowledge surrounding the functional form of the vector field.

\paragraph{Compositional Representation of the Vector Field.}
We represent the vector field of the dynamical system as the composition of a known function -- derived from a priori knowledge -- and a collection of unknown functions that must be learned from data.
\begin{align}
    \dot{x} = h(x,u) = \vectorField (x, u, \unknownTerm_1(\cdot),\hdots, \unknownTerm_\numUnknownTerms (\cdot))
    \label{eq:prior_knowledge-dynamics}
\end{align}
Here, $\vectorField$ is a known differentiable function encoding available prior knowledge on the system's model. 
The functions $\unknownTerm_1, \ldots, \unknownTerm_d$ encode the unknown terms within the underlying model.
The inputs to these functions could themselves be arbitrary functions of the states and control inputs, or even of the outputs of the other unknown terms.

\paragraph{End-To-End Training of the Model.} 
Using the available training data, we learn the unknown functions \(\funk_{\nnParams_1}, \funk_{\nnParams_2}, \ldots, \funk_{\nnParams_{\numUnknownTerms}}\) in an end-to-end fashion.
We parametrize $\unknownTerm_1, \unknownTerm_2, \hdots,\unknownTerm_\numUnknownTerms$ by a collection of neural networks \(\funk_{\nnParams_1}, \funk_{\nnParams_2}, ..., \funk_{\nnParams_{\numUnknownTerms}}\), where \(\nnParams_1, \nnParams_2, \ldots, \nnParams_{\numUnknownTerms}\) are the parameter vectors of each of the individual networks.
For notational simplicity, we define \(\nnParamsAll = (\nnParams_1, \ldots, \nnParams_{\numUnknownTerms})\) and we denote the collection of neural networks \(\funk_{\nnParams_1}, \funk_{\nnParams_2}, ..., \funk_{\nnParams_{\numUnknownTerms}}\) by \(\collectionUnknownTerms\). 

For fixed values of the parameters $\nnParamsAll$, we integrate the current estimate of the dynamics $\vectorField(\cdot, \cdot, \collectionUnknownTerms)$ using a differentiable ODE solver to obtain predictions of the future state. 
More specifically, given an increasing sequence of $\rollout$ distinct points in time $t_i < t_{i+1} < \ldots < t_{i+ \rollout}$, an initial state $\state_i$, and a sequence of control input values $\controlInput_i, \ldots \controlInput_{i+\rollout}$, we use \eqref{eq:ode_integration} to solve for the model-predicted sequence of future states $\state_{i+1}^{\nnParamsAll}, \ldots, \state_{i+\rollout+1}^{\nnParamsAll}$. We assume the control input has a constant value \(\controlInput_{i}\) over the time interval \([t_i, t_{i+1})\).
\begin{equation}
\begin{split}
    \state_{i+1}^\nnParamsAll, \ldots,& \state_{i+\rollout+1}^{\nnParamsAll} = \odesolve(\state_k, \controlInput_{i}, \ldots, \controlInput_{i+\rollout}; \vectorField(\cdot,\cdot,\collectionUnknownTerms))
\end{split} 
\label{eq:ode_integration}
\end{equation}
The loss function \(\loss(\nnParamsAll)\) can then be constructed to minimize the state prediction error over the fixed rollout horizon $\rollout$, as in $\loss(\nnParamsAll) =  \sum_{\trajectory_l \in \dataset} \sum_{(\state_i, \controlInput_i) \in \trajectory_l} \sum_{j=i}^{i+\rollout} \| \state_{j+1}^\nnParamsAll - \state_{j+1}\|^2$, for a given norm.

Finally, assuming that $\vectorField$ is differentiable, we can update the weights in $\nnParamsAll$ using either automatic differentiation, or using the adjoint sensitivity method~\citep{chen2018neural}. 
We note that for each datapoint $(\state_i, \controlInput_i)$ within each trajectory $\trajectory_l$, we use the model to roll out predictions of the next $\rollout$ states, which are in turn used to define the loss function.
This use of rollouts instead of one-step predictions while defining the loss function results in a reduction in error accumulation.
%%%%%%%%%%%%%%%%%%%%%%%%%% CONSTRAINTS
\section{Physics-Informed Neural Network Constraints}

In addition to using a priori physics-based knowledge to dictate the structure of the neural network, we also use such knowledge to derive constraints on the outputs and the internal states of the model. 
More formally, suppose we derive a particular physics-informed model \(\vectorField(\state, \controlInput, \collectionUnknownTerms)\) of the system's vector field with unknown terms parametrized by the collection of neural networks \(\collectionUnknownTerms\).
Recall that \(\nnParamsAll = (\nnParams_1, \nnParams_2, \ldots, \nnParams_{\numUnknownTerms})\) and that by \(\collectionUnknownTerms\) we denote \(\funk_{\nnParams_1}, \ldots, \funk_{\nnParams_{\numUnknownTerms}}\).
Our objective is to solve for a set of parameter values minimizing the loss \(\loss(\nnParamsAll)\) over the training dataset while also satisfying all of the known physics-based constraints.
That is, we aim to solve the optimization problem \eqref{eq:constr_opt_obj}-\eqref{eq:constr_opt_inequal}.
\begin{align}
\min_{\nnParamsAll} \;   \loss(\nnParamsAll) \: \mathrm{s.t.} \;   \equalConstr_i(\state, \controlInput, \collectionUnknownTerms) = 0, \; \forall (\state, \controlInput)\in \constrDomain_{\equalConstr_i}, i \in [\numEqualConstr] , \label{eq:constr_opt_obj}\\
\constr_j(\state, \controlInput, \collectionUnknownTerms) \leq 0, \; \forall (\state, \controlInput) \in \constrDomain_{\inequalConstr_j}, j \in [\numInequalConstr] \label{eq:constr_opt_inequal}
\end{align}
Here, \(\equalConstr_i(\state, \controlInput, \collectionUnknownTerms)\) and \(\inequalConstr_j(\state, \controlInput, \collectionUnknownTerms)\) are differentiable functions capturing the physics-informed equality and inequality constraints respectively. 
We use \([\numEqualConstr] = \{1,\ldots,\numEqualConstr\}\) and \([\numInequalConstr] = \{1,\ldots,\numInequalConstr\}\) to denote the sets indexing these constraints.
For each constraint, we additionally assume there is some sub-set \(\constrDomain_{\equalConstr_i}, \constrDomain_{\inequalConstr_j} \subseteq \stateSpace \times \controlSpace\), also derived from a priori knowledge, over which the constraint should hold true.
As elements \((\state, \controlInput) \in \constrDomain_{\equalConstr_i}, \constrDomain_{\inequalConstr_j} \) are not necessarily included in the trajectory data used for training, the constraints thus provide useful information about potentially unlabeled points.

\paragraph{Enforcing Constraints Throughout the State Space.}
While \eqref{eq:constr_opt_obj}-\eqref{eq:constr_opt_inequal} specifies that each constraint should hold over some known subset of the state space, this formulation leads to a possibly infinite number of constraints.
To approximately solve the problem numerically, we instead select a finite set \(\collocationPoints = \{(\state_1, \controlInput_1), (\state_2, \controlInput_2), \ldots (\state_{\numCollocationPoints}, \controlInput_{\numCollocationPoints})\} \subseteq \stateSpace \times \controlSpace\) of points at which we enforce the constraints.
That is, we replace the possibly infinite sets of constraints given by \eqref{eq:constr_opt_obj}-\eqref{eq:constr_opt_inequal} with the finite set of constraints given by \eqref{eq:constr_opt_equal_collocation}--\eqref{eq:constr_opt_inequal_collocation}.
\begin{align}
    &  \equalConstr_i(\state, \controlInput, \collectionUnknownTerms) = 0, \forall (\state, \controlInput)\in \collocationPoints \cap \constrDomain_{\equalConstr_i}, i \in [\numEqualConstr]  \label{eq:constr_opt_equal_collocation}  \\
    & \constr_i(\state, \controlInput, \collectionUnknownTerms) \leq 0, \forall (\state, \controlInput) \in \collocationPoints \cap \constrDomain_{\inequalConstr_j}, j \in [\numInequalConstr] \label{eq:constr_opt_inequal_collocation}
\end{align}
Intuitively, by optimizing the loss function $\loss(\nnParamsAll)$ in \eqref{eq:constr_opt_obj} subject to \eqref{eq:constr_opt_equal_collocation}--\eqref{eq:constr_opt_inequal_collocation}, we are finding the solution that fits the training data as well as possible, while also satisfying all of the relevant constraints at each point within a finite set \(\collocationPoints\) of representative states throughout \(\stateSpace \times \controlSpace\).

\paragraph{The Augmented Lagrangian Method.}
In order to solve this optimization problem, we use a stochastic gradient descent (SGD) variant of the augmented Lagrangian method, as proposed by \citep{lu2021physics, toussaint2014novel} for the numerical solution of constrained optimization problems.
For each constraint in \eqref{eq:constr_opt_equal_collocation}-\eqref{eq:constr_opt_inequal_collocation}, we define a separate Lagrange variable.
That is, we define an individual variable \(\lagrangeVar^{\mathrm{eq}}_{i,k}\) for each equality constraint \(\equalConstr_i, i \in [\numEqualConstr]\) evaluated at each point \((\state_k, \controlInput_k) \in \collocationPoints \cap \constrDomain_{\equalConstr_i}\), and similarly \(\lagrangeVar^{\mathrm{ineq}}_{j,k}\) at each point  \((\state_k, \controlInput_k) \in \collocationPoints \cap \constrDomain_{\inequalConstr_j}\) for all the inequality constraints \(\inequalConstr_j\).
The augmented Lagrangian of the optimization problem is given by \eqref{eq:augmented_lagrangian}.

\begin{align}
    \augmentedLagrangian&(\nnParamsAll, \constrPenalty, \lambda) = \loss(\nnParamsAll) + \sum_{\substack{i\in [\numEqualConstr] \\ (\state_k, \controlInput_k) \in \collocationPoints \cap \constrDomain_{\equalConstr_i}}} \constrPenalty \equalConstr_i(\state_k, \controlInput_k, \collectionUnknownTerms)^2  + \sum_{\substack{i\in [\numEqualConstr] \\ (\state_k, \controlInput_k) \in \collocationPoints \cap \constrDomain_{\equalConstr_i}}} \lambda_{i,k}^{\mathrm{eq}} \equalConstr_i(\state_i, \controlInput_i, \collectionUnknownTerms) \label{eq:augmented_lagrangian}\\
    & + \sum_{\substack{j\in [\numInequalConstr] \\ (\state_k, \controlInput_k)\in \collocationPoints \cap \constrDomain_{\inequalConstr_j}}} \constrPenalty [\lambda_{j,k}^{ineq} > 0 \lor \inequalConstr_j > 0] \inequalConstr_j(\state_k, \controlInput_k, \collectionUnknownTerms)^2 + \sum_{\substack{j\in [\numInequalConstr] \\ (\state_k, \controlInput_k) \in \collocationPoints \cap \constrDomain_{\inequalConstr_j}}} \lambda_{j,k}^{\mathrm{ineq}} \inequalConstr_j(\state_k, \controlInput_k, \collectionUnknownTerms) \nonumber
\end{align}
\paragraph{The Proposed Training Algorithm.}
Algorithm \ref{alg:algorithm} outlines the proposed approach to optimizing the loss function $\loss(\nnParamsAll)$ in \eqref{eq:constr_opt_obj} subject to \eqref{eq:constr_opt_equal_collocation}--\eqref{eq:constr_opt_inequal_collocation} via the augmented Lagrangian method.
In the algorithm, we use the notation \(x_{+} = \max\{0,x\}\).
We initialize values for \(\constrPenalty\) and \(\lambda\) and minimize \(\augmentedLagrangian(\nnParamsAll, \constrPenalty, \lambda)\) via SGD over \(\nnParamsAll\) while holding the values of \(\constrPenalty\) and \(\lambda\) fixed.
To prevent the gradient descent from getting stuck in local minima, Algorithm~\ref{alg:algorithm} randomly samples a subset of points from \(\collocationPoints\) at each gradient update, instead of including the entire set in the definition of \(\augmentedLagrangian(\nnParamsAll, \constrPenalty, \lambda)\).
This random selection of points at which to evaluate the constraint violations is akin to the random sampling of minibatches of training data during traditional SGD.
Once this inner loop SGD has converged, we update the values of \(\constrPenalty\) and \(\lambda\) according to the update rules outlined in Algorithm \ref{alg:algorithm}.
This process is repeated until the constraints are all satisfied, to within an allowed tolerance value \(\constrTol\).

\begin{algorithm}[H]
\caption{Training Algorithm}
\label{alg:algorithm}
\textbf{Input}: $\vectorField(\cdot)$, $\{\equalConstr_i(\cdot)\}_{i \in [\numEqualConstr]}$, $\{\inequalConstr_j(\cdot)\}_{j \in [\numInequalConstr]}$, $\dataset$, $\collocationPoints$\\
\textbf{Parameter}: $\constrTol$, $\constrPenalty_0$, $\constrPenalty_{mult}$, $N_{\mathrm{trainBatch}}$, $N_{\mathrm{constrBatch}}$\\
\textbf{Output}: Model parameters $\nnParamsAll$.

\begin{algorithmic}[1] %[1] enables line numbers

\STATE Initialize parameters \(\nnParamsAll\); $\lagrangeVar^\mathrm{eq}_{i,k} \gets 0$; $\lagrangeVar^\mathrm{ineq}_{j,k} \gets 0$; $\constrPenalty \gets \constrPenalty_0$.

\WHILE{$\sum_{i} \sum_{k} |\equalConstr_i(x_k,u_k,\collectionUnknownTerms)| + \sum_{j} \sum_{k} (\inequalConstr_j(x_k,u_k,\collectionUnknownTerms))_+ \geq \constrTol$} 

    \WHILE{\textbf{not} SGDStoppingCriterion()}
        \STATE $\dataset_{batch} \gets Sample(\dataset, N_{trainBatch})$; $\collocationPoints_{batch} \gets Sample(\collocationPoints, N_{constrBatch})$
        \STATE $\nnParamsAll \gets optimUpdate(\mathcal{L}, \nnParamsAll, \dataset_{batch}, \collocationPoints_{batch})$
    \ENDWHILE

    \STATE $\lagrangeVar^\mathrm{eq}_{i,k} \gets \lagrangeVar^\mathrm{eq}_{i,k} + 2 * \constrPenalty * \equalConstr_i(\state_k, \controlInput_k, \collectionUnknownTerms)$; $\lagrangeVar^\mathrm{ineq}_{j,k} \gets (\lagrangeVar^\mathrm{ineq}_{j,k} + 2 * \constrPenalty * \inequalConstr_j(\state_k, \controlInput_k, \collectionUnknownTerms))_+$
    \STATE $\constrPenalty \gets \constrPenalty * \constrPenalty_{mult}$
\ENDWHILE

\end{algorithmic}

\end{algorithm}
\section{Experimental Results}

\subsection{Learning the Dynamics of a Double Pendulum}
\begin{figure}
    \centering
    \input{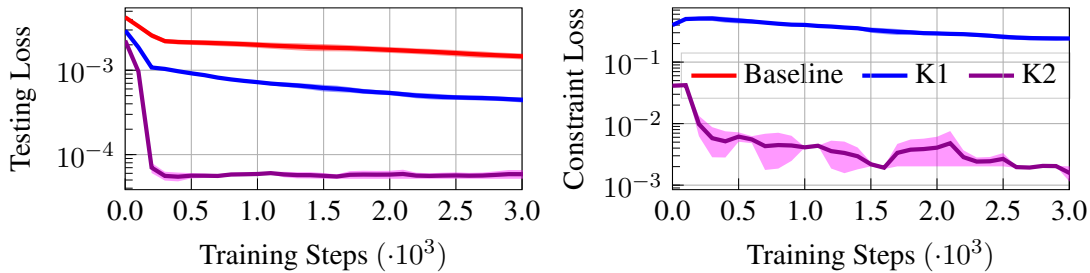}
    \vspace{-3mm}
    \caption{
    Double pendulum numerical results.
    A single trajectory that is 3 seconds in length is used as training data.
    The shaded areas illustrate the variance of the results over five runs.}
    \vspace{-5mm}
    \label{fig:dp_results_fig}
\end{figure}
The dynamics of the double pendulum are described by \eqref{eq:double_pend_dynamics}.
\begin{equation}
    \Ddot{\phi_1} = (\funk_{1} - \alpha_1 \funk_{2}) / (1 - \alpha_1 \alpha_2); \;\;  \Ddot{\phi_2} = (- \alpha_2 \funk_{1} + \funk_{2})/(1 - \alpha_1 \alpha_2)
    \label{eq:double_pend_dynamics}
\end{equation}

Here, \(\phi_1\) and \(\phi_2\) specify the angular position of the first and second links of the pendulum, \(\alpha_1(\phi_1, \phi_2) \propto cos(\phi_1 - \phi_2)\), \(\alpha_2(\phi_1, \phi_2) \propto cos(\phi_2 - \phi_1)\), and \(\funk_1(\phi_1, \phi_2, \dot{\phi}_1, \dot{\phi}_2)\), \(\funk_2(\phi_1, \phi_2, \dot{\phi}_1, \dot{\phi}_2)\) are both complicated trigonometric functions of the state variables.
By setting \(\state_1 = \phi_1\), \(\state_2 = \phi_2\), \(\state_3 = \dot{\phi}_1\), and \(\state_4 = \dot{\phi}_2\) we can thus express the dynamics as \(\dot{\state} = h(\state)\), where \(h_1(\state) = \state_3\), \(h_2(\state) = \state_4\), and \(h_3(\state)\), \(h_4(\state)\) are both given by \eqref{eq:double_pend_dynamics}.
The full system dynamics are provided in Appendix \ref{sec:supp_mat_double_pend_details}.

\paragraph{Defining a Baseline for Comparison.} 
As a baseline, we parametrize the vector field \(h\) by a single neural network \(\funk_{\nnParams_1}\).
In terms of the framework outlined in the previous sections, we thus have \(\nnParamsAll = \nnParams_1\) and \(\vectorField(\state, \collectionUnknownTerms) = \funk_{\nnParams_1}(\state)\).

\paragraph{(K1) Incorporating Knowledge of Equation \eqref{eq:double_pend_dynamics}.}
In order to demonstrate how arbitrary knowledge surrounding the system's vector field can be incorporated into the dynamics model, we assume that equation \eqref{eq:double_pend_dynamics} is known, along with \(\alpha_1(\cdot)\) and \(\alpha_2(\cdot)\). 
However, we assume \(\funk_1(\cdot)\) and \(\funk_2(\cdot)\) are unknown.
We parametrize \(\funk_1\) and \(\funk_2\) by two neural networks \(\funk_{\nnParams_1}\) and \(\funk_{\nnParams_2}\), respectively.
The proposed model for the vector field may thus be expressed as \(\vectorField_1(\state, \collectionUnknownTerms) = \state_3\), \(\vectorField_2(\state, \collectionUnknownTerms) = \state_4\), and with \(\vectorField_3(\state, \collectionUnknownTerms)\), \(\vectorField_4(\state, \collectionUnknownTerms)\) both being functions of \(\alpha_1\), \(\alpha_2\), \(\funk_{\nnParams_1}\), \(\funk_{\nnParams_2}\), as in equation \eqref{eq:double_pend_dynamics}.

\paragraph{(K2) Incorporating Symmetry Constraints on \(\funk_{1}\) and \(\funk_{2}\).}
To demonstrate that the proposed training algorithm properly enforces constraints on the system's dynamics, we impose equality constraints on \(\funk_{\nnParams_1}\), \(\funk_{\nnParams_2}\) derived from symmetries in the vector field. 
In particular, we impose four separate equality constraints, an example of which is as follows: \(\funk_1(\state_{1:2}, \state_{3:4}) = -\funk_1(-\state_{1:2}, \state_{3:4})\).
These equality constraints are enforced at a collection of points \(\collocationPoints\) that are sampled uniformly throughout the state space.
\paragraph{Experimental Setup.}
Each neural network \(\funk_{\nnParams_i}(\state)\) is a multilayer perceptron (MLP) with ReLu activation functions and two hidden layers with 128 nodes each. 
We use a rollout horizon of \(\rollout=5\) when defining the loss function and we randomly sample \(|\collocationPoints| = 10000\) points throughout the state space \(\stateSpace\) at which we enforce the above equality constraints. 
Further details on the experiment implementation and hyperparameters are provided in Appendix \ref{sec:supp_mat_implementation_details}.

\paragraph{Experimental Results.}
Figure \ref{fig:dp_results_fig} plots the results of training the various physics-informed models using only a single trajectory of data, which equivalent to 3 seconds worth of observations.
We observe that in terms of prediction losses, \((K1)\) performs better than the baseline, and \((K2)\) performs better than both the baseline and \((K1)\); its testing loss at convergence is roughly an order of magnitude lower than that of \((K1)\).
Furthermore, we observe that the constraint loss of \((K2)\) -- which measures the average value of the model's violation of the equality constraints -- is roughly two orders of magnitude lower than that of \((K1)\).
These results demonstrate the proposed framework's effectiveness at incorporating a priori system knowledge into the dynamics model.
As more detailed knowledge is included into the model, the testing loss at convergence is significantly reduced. 
Furthermore, the gap in constraint loss between \((K1)\) and \((K2)\) demonstrates the effectiveness of the proposed framework in learning to fit the training data while simultaneously enforcing constraints.

%%%%%%%%%%%%%%%%%% BRAX RESULTS

\subsection{Learning the Dynamics of Controlled, Rigid, Multi-Body Systems}

\begin{figure*}[t]
    \centering
    \input{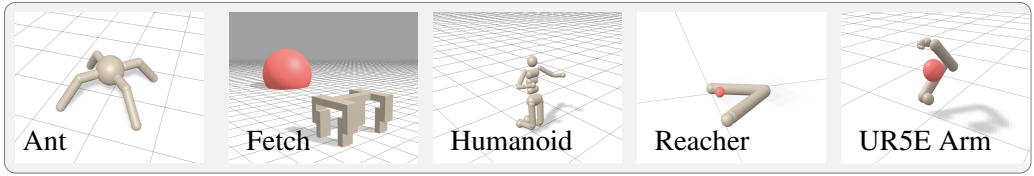}
    \vspace{-3mm}
    \caption{
    Illustration of the suite of simulated robotic systems used during testing.
    }
    \label{fig:environment_suite}
\end{figure*}

In this section we apply the proposed framework to a suite of robotic systems simulated in Brax \citep{freeman2021brax}. 
These systems feature control inputs, non-linear dynamics, and external forces.
Figure \ref{fig:environment_suite} illustrates the specific robotic systems we use for testing.

\paragraph{The Governing Equations of Motion.} The dynamics of all of the environments listed in Figure \ref{fig:environment_suite} can be described by the general equations of motion given in \eqref{eq:eq_of_motion}.
\begin{equation}
    \Ddot{q} = \massMat(q)^{-1}[\corForce(q,\dot{q}) + \actuationForce(q, \dot{q}, \controlInput) + \jacobian(q)\mathcal{\contactForce}(q, \dot{q}, \controlInput)] 
    \label{eq:eq_of_motion}
\end{equation}
Here, \(q\) and \(\dot{q}\) represent the system state and its time derivative respectively, $\massMat(q)$ represents the system's mass matrix, $\corForce(q, \dot{q})$ is a vector representing the Coriolis forces, $\actuationForce(q, \dot{q}, \controlInput)$ is a vector representing the actuation forces (given the control input $\controlInput$), \(\jacobian(q)\) is the system's Jacobian matrix, and \(\contactForce(q, \dot{q}, \controlInput)\) is a vector representing the contact forces.
By setting \(\state_1 = q\) and \(\state_2 = \dot{q}\), we may re-write the system in the form \(\dot{x} = h(\state, \controlInput)\).

\paragraph{Defining a Baseline for Comparison.} 
Similarly to the double pendulum baseline, we use a single neural network to parametrize the vector field, i.e. \(\vectorField(\state, \controlInput, \collectionUnknownTerms) = \funk_{\nnParams_1}(\state, \controlInput)\).

\paragraph{(K1) Incorporating Basic Knowledge on the Vector Field.}
A relatively simple piece of a priori knowledge is that \(\dot{\state}_1 = \state_2\).
This follows directly from our formulation of the system's dynamics, and it provides a useful piece of inductive bias surrounding the structure of the vector field.
In our proposed framework, we again parametrize the vector field using a single neural network \(\funk_{\nnParams_1}\), however, we now have \(\vectorField_1(\state, \controlInput, \collectionUnknownTerms) = \state_2\) and \(\vectorField_2(\state, \controlInput, \collectionUnknownTerms) = \funk_{\nnParams_1}(\state, \controlInput)\).

\paragraph{(K2) Incorporating Knowledge of the Mass Matrix.}
We now assume the mass matrix \(\massMat(\state_1)\) and the Coriolis force \(\corForce(\state_1, \state_2)\) terms are known a priori.
To use this a priori knowledge, we parametrize the actuation force term and the contact force term using individual neural networks \(\funk_{\nnParams_1}\) and \(\funk_{\nnParams_2}\).
Our structured neural network model of the system dynamics may be written as follows: \(\vectorField_1(\state, \controlInput, \collectionUnknownTerms) = \state_2\) and \(\vectorField_2(\state, \controlInput, \collectionUnknownTerms) = \massMat^{-1}(\state_1)\corForce(\state_1, \state_2) + \funk_{\nnParams_1}(\state, \controlInput) + \funk_{\nnParams_2}(\state, \controlInput)\).
\begin{figure}[t]
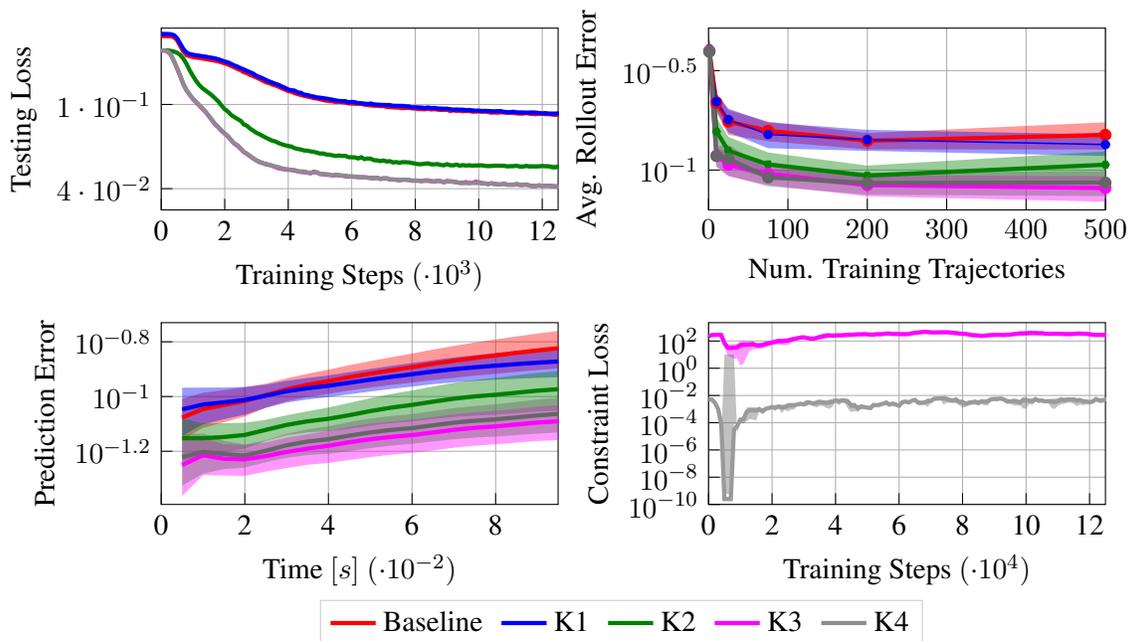

    \centering
    % This file was created with tikzplotlib v0.9.12.
\begin{tikzpicture}

\definecolor{color0}{rgb}{1,0,1}

\begin{groupplot}[group style={group name = plots, group size=2 by 2, vertical sep=1.5cm, horizontal sep=2.0cm}]

%%%%%%%%%%%%%% testing loss vs training steps plot
\nextgroupplot[
width=0.45\textwidth, 
height=4cm,
legend cell align={left},
legend columns=5,
legend style={
    anchor=south west,
    at={(0.4, -2.4)},
    fill opacity=0.8, 
    draw opacity=1, 
    text opacity=1,
    draw=white!80!black, 
    /tikz/every even column/.append style={column sep=0.15cm}
},
log basis y={10},
% scaled x ticks=manual:{}{\pgfmathparse{#1}},
tick align=inside,
tick pos=left,
x grid style={white!69.0196078431373!black},
xlabel={Training Steps \((\cdot 10^3)\)},
xmajorgrids,
xmin=0, 
xmax=12500,
xtick={0,2000,4000,6000,8000,10000,12000},
xticklabels={0, 2, 4, 6, 8, 10, 12},
xtick style={color=black},
scaled x ticks=false,
y grid style={white!69.0196078431373!black},
ylabel={Testing Loss},
ymajorgrids,
ymin=0.0317490094894555, ymax=0.229388645238984,
ytick={0.04, 0.1},
yticklabels={\(4 \cdot 10^{-2}\), \(1 \cdot 10^{-1}\)},
ymode=log,
ytick style={color=black},
]
\input{plots/fetch/train_test_loss_data}

%%%%%%%%%%%%%%% rollout err vs num training trajectories plot
\nextgroupplot[
width=0.45\textwidth, 
height=4cm,
legend cell align={left},
legend cell align={left},
legend style={fill opacity=0.8, draw opacity=1, text opacity=1, draw=white!80!black},
log basis y={10},
tick align=inside,
tick pos=left,
x grid style={white!69.0196078431373!black},
xlabel={Num. Training Trajectories},
xmajorgrids,
xmin=-0, xmax=500,
xtick style={color=black},
y grid style={white!69.0196078431373!black},
ylabel={Avg. Rollout Error},
ymajorgrids,
ymin=0.06299942014023, ymax=0.517269287287668,
ymode=log,
ytick style={color=black}
]
\input{plots/fetch/traj_err_data}

%%%%%%%%%% Prediction error plot
\nextgroupplot[
width=0.45\textwidth, 
height=4cm,
legend cell align={left},
legend style={
  fill opacity=0.8,
  draw opacity=1,
  text opacity=1,
  at={(0.03,0.97)},
  anchor=north west,
  draw=white!80!black
},
log basis y={10},
tick align=inside,
tick pos=left,
x grid style={white!69.0196078431373!black},
xlabel={Time \([s]\) \((\cdot 10^{-2})\)},
xmajorgrids,
xmin=0.0, xmax=0.09494999977760017,
scaled x ticks=false,
xtick={0, 0.02, 0.04, 0.06, 0.08},
xticklabels={0, 2, 4, 6, 8},
xtick style={color=black},
y grid style={white!69.0196078431373!black},
ylabel={Prediction Error},
ymajorgrids,
ymin=0.0403423932359269, ymax=0.186214482340161,
ymode=log,
ytick style={color=black}
]
\input{plots/fetch/rel_err_data}

%%%%%%%% CONSTRAINT LOSS PLOT
\nextgroupplot[
width=0.45\textwidth, 
height=4cm,
legend cell align={left},
legend style={fill opacity=0.8, draw opacity=1, text opacity=1, draw=white!80!black},
log basis y={10},
tick align=inside,
tick pos=left,
unbounded coords=jump,
x grid style={white!69.0196078431373!black},
xlabel={Training Steps \((\cdot 10^{4})\)},
xmin=0, 
xmax=12500,
xtick={0,2000,4000,6000,8000,10000,12000},
xticklabels={0, 2, 4, 6, 8, 10, 12},
xtick style={color=black},
scaled x ticks=false,
y grid style={white!69.0196078431373!black},
ylabel={Constraint Loss},
ymin=1e-10, ymax=2225.0296872671,
xmajorgrids,
ymajorgrids,
ymode=log,
ytick style={color=black},
ytick={1e-10,1e-08,1e-06,0.0001,0.01,1,100,10000},
yticklabels={
  \(\displaystyle {10^{-10}}\),
  \(\displaystyle {10^{-8}}\),
  \(\displaystyle {10^{-6}}\),
  \(\displaystyle {10^{-4}}\),
  \(\displaystyle {10^{-2}}\),
  \(\displaystyle {10^{0}}\),
  \(\displaystyle {10^{2}}\),
  \(\displaystyle {10^{4}}\)
}
]
\input{plots/fetch/constraint_loss_data}
\end{groupplot}

\end{tikzpicture}
    \vspace*{-6.5mm}
    \caption{Numerical results for the \textit{Fetch} robotic system.}
    \label{fig:fetch_results_si2}
\end{figure}
\paragraph{(K3) Incorporating Knowledge of the Jacobian Matrix.} 
We assume, in addition to the a priori knowledge assumed in \((K3)\), that the system's Jacobian matrix \(\jacobian(\state_1)\) is also known:
\(\vectorField_1(\state, \controlInput, \collectionUnknownTerms) = \state_2\), \(\vectorField_2(\state, \controlInput, \collectionUnknownTerms) = \massMat^{-1}(\state_1)\corForce(\state, \state_2) + \funk_{\nnParams_1} + \jacobian(\state_1)\funk_{\nnParams_2}\).
\paragraph{(K4) Incorporating Contact Force Constraints.} 
We use the laws of friction to  derive constraints on the contact forces between the robotic systems and the ground.
In particular, we have \(\contactForce_{norm}(\state, \controlInput) \leq 0\) and \(||\contactForce_{tang}(\state, \controlInput)|| \geq \mu \contactForce_{norm}(\state, \controlInput)\), where \(\contactForce_{norm}\) represents the normal component of the contact forces and \(\contactForce_{tang}\) is the tangential force vector.
We assume \((K4)\) has access to the same amount of a priori knowledge as \((K3)\), and that it additionally imposes the above inequality constraints on the neural network \(\funk_{\nnParams_2}\), which represents the contact force term.

\paragraph{Experimental Setup.} 
We provide the experimental results for the \textit{Fetch} robotic system.
This robot has a 143-dimensional state space and a 10-dimensional control input space.
To parametrize each unknown term \(\funk_{\nnParams_i}\) we use a multilayer perceptron (MLP) with ReLu activation functions and two hidden layers of 256 nodes each.
We use a rollout horizon of \(\rollout = 5\) to define the loss function.
Further details, as well as the experimental results for the other robotic systems, are provided in Appendix \ref{sec:supp_mat_robotics_details}.

\paragraph{Experimental Results.} 
Figure \ref{fig:fetch_results_si2} illustrates the results.
Similarly to the double pendulum study, we observe from Figure \ref{fig:fetch_results_si2} that as we increase the amount of a priori knowledge incorporated into the model, the average prediction error correspondingly decreases.
In particular, the \textit{Avg. Rollout Error} plot shows that by incorporating a priori knowledge of the mass matrix, \((K2)\) consistently makes more accurate predictions than either the baseline or \((K1)\), regardless of the number of trajectories included in the training dataset.
From the \textit{Constraint Loss} plot we observe a gap of four orders of magnitude between the constraint loss of \((K3)\) and \((K4)\): the proposed algorithm learns a model that effectively enforces the contact force constraints incorporated into \((K4)\). 
\section{Related Work}
\label{sec:related_work}

Techniques for \textit{nonlinear system identification} -- the construction of mathematical models of dynamical systems from empirical observations -- often rely on a priori knowledge of a suitable collection of nonlinear basis functions \citep{brunton2016discovering, brunton2016sparse, williams2014kernel, williams2015data}.
However, choosing such sets of basis functions can be challenging, particularly for systems with complex dynamics and high-dimensional state spaces.
\cite{li2017extended, takeishi2017learning, champion2019data, yeung2019learning} thus use neural networks to learn appropriate collections of nonlinear functions to be used in conjunction with these techniques.
More closely related to our work, \cite{raissi2018multistep, lu2018beyond, dupont2019augmented, kelly2020learning, massaroli2020dissecting} use neural networks to represent differential equations.
\cite{chen2018neural} introduce the neural ODE; neural networks parametrizing ordinary differential equations.
However, the above works do not consider how physics-based knowledge might be incorporated into neural networks, nor how physics-based constraints can be enforced during training. These are the primary focuses of this paper.

The inclusion of physics-based knowledge in the training of neural networks has, however, been studied extensively over the past several years. 
In particular, \cite{lu2021physics, raissi2019physics, raissi2018deep, han2018solving, sirignano2018dgm, long2018pde, long2019pde} use neural networks for the solution and discovery of partial differential equations (PDE).
Our work instead focuses on using neural networks to parametrize the vector field of dynamical systems.

More closely related to our work, a collection of recent papers study how physics-based knowledge can be incorporated into neural networks for the learning of dynamics models from trajectory data \citep{zhong2021benchmarking, hernandez2021structure, rackauckas2020universal}. 
Many of these works either use the Lagrangian or the Hamiltonian formulation of dynamics to inform the structure of a neural ODE, as in \cite{cranmer2020lagrangian, lutter2019deep, roehrl2020modeling, zhong2020unsupervised, allen2020lagnetvip} vs. \cite{greydanus2019hamiltonian, matsubara2019deep, toth2019hamiltonian}.
\cite{finzi2020simplifying} use the method of Lagrange multipliers to explicitly enforce holonomic constraints on the state variables in Lagrangian and Hamiltonian neural networks.
A number of works have also studied physics-informed neural ODEs in the context of learning control-oriented dynamics models \citep{zhong2019symplectic, zhong2020dissipative, roehrl2020modeling, duong2021hamiltonian, gupta2020structured, menda2019structured, zhong2021differentiable, shi2019neural}.
We note however, that the majority of the above works employ a specific neural ODE structure, rendering them less flexible than the general framework that we propose. 
Furthermore, the examples explored in these works typically feature low-dimensional state spaces that often do not include energy dissipation or external forces.
By contrast, we apply the proposed approach to a variety of high-dimensional and complex systems.
Finally, the above works do not enforce general physics-based constraints on the dynamics model.

A number of other recent works have also studied imposing constraints on the outputs of deep neural networks \citep{marquez2017imposing, fioretto2020lagrangian, nandwani2019primal, kervadec2020constrained}. 
Similarly to our work, \cite{lu2021physics, dener2020training} consider an augmented Lagrangian approach to the enforcement of physics-based constraints.
However, they do so in different contexts than ours; the former focuses on PDE-constrained inverse design problems while the latter uses neural networks to approximate the Focker-Planck operator used in fusion simulations.
\section{Conclusions}

In this work we present a framework for the incorporation of a priori physics-based knowledge into neural network models of dynamical systems.
This physics based knowledge is incorporated either by influencing the structure of the neural network, or as constraints on the model's outputs and internal states.
We present a suite of numerical experiments that exemplify the effectiveness of the proposed approach: the inclusion of increasingly detailed forms of side-information leads to increasingly accurate model predictions. 
We also demonstrate that the framework effectively enforces physics-based constraints on the outputs and internal states of the models.
Future work will aim to use these physics-informed dynamics models for control.

% Acknowledgments---Will not appear in anonymized version
\acks{This work was supported in part by ONR N00014-22-1-2254, AFOSR FA9550-19-1-0005, and \\NSF 1646522.}

\bibliography{bibliography.bib}

%%%%%%%%%% Supplementary material

\newpage

\onecolumn

\begin{center} % supp mat title
    \begin{LARGE}
        \textbf{Neural Networks with Physics-Informed Architectures and Constraints for Dynamical Systems Modeling: Supplementary Material}
    \end{LARGE}
\end{center}

\appendix
\section{Implementation Details}
\label{sec:supp_mat_implementation_details}
All numerical experiments were implemented using the python library \textit{Jax} \cite{jax2018github}, in order to take advantage of its automatic differentiation and just-in-time compilation features.
Project code is available at \href{https://github.com/wuwushrek/physics_constrained_nn}{https://github.com/wuwushrek/physics\_constrained\_nn}.

All experiments were run locally on a desktop computer with an Intel i9-9900 3.1 GHz CPU with 32 GB of RAM and a GeForce RTX 2060, TU106. A complete training run of roughly $10000$ gradient iterations, takes approximately $5$ to $10$ minutes of wall-clock time on the GPU.

\paragraph{Hyperparameters.} 
Table \ref{tab:hyperparam_all} lists the values of the hyperparameters that were used to generate the experimental results. 

\begin{table}
\centering
    {
        \begin{tabu}{llllllll}
            \hline
            
            & & \multicolumn{1}{c}{\pbox{15cm}{Double\\pendulum}} & \multicolumn{1}{c}{Ant} & \multicolumn{1}{c}{Fetch} & \multicolumn{1}{c}{Humanoid} & \multicolumn{1}{c}{Reacher} & \multicolumn{1}{c}{UR5E}  \\
        
            \hline
            
            \multirow{3}{*}{\pbox{15cm}{MLP \\ Construction}} 
            & \multicolumn{1}{l}{\pbox{10cm}{Hidden \\ layers}} & \multicolumn{1}{l}{2} & \multicolumn{1}{l}{2} & \multicolumn{1}{l}{2} &  \multicolumn{1}{l}{2} & \multicolumn{1}{l}{2} & \multicolumn{1}{l}{2}\\\cline{2-8}
            & \multicolumn{1}{l}{\pbox{7cm}{Nodes \\ per layer}} & \multicolumn{1}{l}{256} & \multicolumn{1}{l}{512} & \multicolumn{1}{l}{512} &  \multicolumn{1}{l}{512} & \multicolumn{1}{l}{256} & \multicolumn{1}{l}{512} \\\cline{2-8}
            & \multicolumn{1}{l}{\pbox{7cm}{Activation \\ function}} & \multicolumn{1}{l}{ReLu} & \multicolumn{1}{l}{ReLu} &  \multicolumn{1}{l}{ReLu} & \multicolumn{1}{l}{ReLu} & \multicolumn{1}{l}{ReLu} & \multicolumn{1}{l}{ReLu} \\
            
            \hline
            
            \multirow{3}{*}{\pbox{15cm}{Training \\ Parameters}}
            & \multicolumn{1}{l}{Optimizer} & \multicolumn{1}{l}{ADAM} & \multicolumn{1}{l}{ADAM} & \multicolumn{1}{l}{ADAM} &  \multicolumn{1}{l}{ADAM} & \multicolumn{1}{l}{ADAM} & \multicolumn{1}{l}{ADAM}\\\cline{2-8}
            & \multicolumn{1}{l}{Learning rate} & \multicolumn{1}{l}{\(5e-3\)} & \multicolumn{1}{l}{\(5e-3\)} & \multicolumn{1}{l}{\(1e-3\)} &  \multicolumn{1}{l}{\(1e-3\)} & \multicolumn{1}{l}{\(1e-2\)} & \multicolumn{1}{l}{\(1e-2\)}\\\cline{2-8}
            & \multicolumn{1}{l}{Minibatch size} & \multicolumn{1}{l}{64} & \multicolumn{1}{l}{64} & \multicolumn{1}{l}{64} &  \multicolumn{1}{l}{64} & \multicolumn{1}{l}{64} & \multicolumn{1}{l}{64} \\
            \cline{2-8}
            & \multicolumn{1}{l}{\pbox{7cm}{Early stopping \\ patience}} & \multicolumn{1}{l}{1000} & \multicolumn{1}{l}{1000} & \multicolumn{1}{l}{1000} &  \multicolumn{1}{l}{1000} & \multicolumn{1}{l}{1000} & \multicolumn{1}{l}{1000} \\ \cline{2-8}
            
            \hline
            
            \multirow{6}{*}{\pbox{15cm}{Constraint \\ Enforcement}}
            & \multicolumn{1}{l}{Num. points\(\numCollocationPoints\)} & \multicolumn{1}{l}{\(10,000\)} & \multicolumn{1}{l}{\(50,000\)} & \multicolumn{1}{l}{\(60,000\)} &  \multicolumn{1}{l}{\(120,000\)} & \multicolumn{1}{l}{\(-\)} & \multicolumn{1}{l}{\(-\)}\\\cline{2-8}
            & \multicolumn{1}{l}{\pbox{7cm}{Minibatch size \\ \(|\numCollocationPoints_{batch}|\)}} & \multicolumn{1}{l}{\(256\)} & \multicolumn{1}{l}{\(128\)} & \multicolumn{1}{l}{\(128\)} &  \multicolumn{1}{l}{\(128\)} & \multicolumn{1}{l}{\(-\)} & \multicolumn{1}{l}{\(-\)} \\ \cline{2-8}
            & \multicolumn{1}{l}{\pbox{7cm}{Initial penalty \\ term \(\constrPenalty_0\)}} & \multicolumn{1}{l}{\(1e-3\)} & \multicolumn{1}{l}{\(1e-3\)} & \multicolumn{1}{l}{\(1e-3\)} &  \multicolumn{1}{l}{\(1e-3\)} & \multicolumn{1}{l}{\(-\)} & \multicolumn{1}{l}{\(-\)} \\ \cline{2-8}
            & \multicolumn{1}{l}{\pbox{7cm}{Penalty \\ multiplier \(\constrPenalty_{mult}\)}} & \multicolumn{1}{l}{\(1.5\)} & \multicolumn{1}{l}{\(1.5\)} & \multicolumn{1}{l}{\(1.5\)} &  \multicolumn{1}{l}{\(1.5\)} & \multicolumn{1}{l}{\(-\)} & \multicolumn{1}{l}{\(-\)} \\ \cline{2-8}
            & \multicolumn{1}{l}{\pbox{7cm}{Constraint loss \\ tolerance \(\constrTol\)}} & \multicolumn{1}{l}{\(1e-4\)} & \multicolumn{1}{l}{\(5e-4\)} & \multicolumn{1}{l}{\(1e-3\)} &  \multicolumn{1}{l}{\(1e-3\)} & \multicolumn{1}{l}{\(-\)} & \multicolumn{1}{l}{\(-\)} \\ \cline{2-8}
            
            \hline
            
            \multirow{2}{*}{\pbox{15cm}{Other \\ Parameters}}
            & \multicolumn{1}{l}{\pbox{7cm}{Rollout \\ length \(\rollout\)}} & \multicolumn{1}{l}{5} & \multicolumn{1}{l}{5} & \multicolumn{1}{l}{4} &  \multicolumn{1}{l}{4} & \multicolumn{1}{l}{8} & \multicolumn{1}{l}{8} \\ \cline{2-8}
            & \multicolumn{1}{l}{\pbox{7cm}{Numumber of \\ random seeds}} & \multicolumn{1}{l}{3} & \multicolumn{1}{l}{3} & \multicolumn{1}{l}{3} &  \multicolumn{1}{l}{3} & \multicolumn{1}{l}{3} & \multicolumn{1}{l}{3}\\ \cline{2-8}
            
            \hline
            
        \end{tabu}
        }
    \caption{The hyperparameter values used in the experiments.}
    \label{tab:hyperparam_all}
\end{table}

\newpage
\section{Supplementary Double Pendulum Details and Results}
\label{sec:supp_mat_double_pend_details}

\paragraph{Dynamics of Double Pendulum.}
The full dynamics of the double pendulum are given by \eqref{eq:double_pend_dynamics2}-\eqref{eq:g2_equation}.

\begin{equation}
    \Ddot{\phi_1} = \frac{\funk_{1}(\phi_1, \phi_2, \dot{\phi_1}, \dot{\phi_2}) - \alpha_1(\phi_1, \phi_2) \funk_{2}(\phi_1, \phi_2, \dot{\phi_1}, \dot{\phi_2})}{1 - \alpha_1(\phi_1, \phi_2) \alpha_2(\phi_1, \phi_2)} \label{eq:double_pend_dynamics2}
\end{equation}

\begin{equation}
    \Ddot{\phi_2} = \frac{- \alpha_2(\phi_1, \phi_2) \funk_{1}(\phi_1, \phi_2, \dot{\phi_1}, \dot{\phi_2}) + \funk_{2}(\phi_1, \phi_2, \dot{\phi_1}, \dot{\phi_2})}{1 - \alpha_1(\phi_1, \phi_2) \alpha_2(\phi_1, \phi_2)}
\end{equation}

\begin{equation}
    \alpha_1(\phi_1, \phi_2) = \frac{l_1}{l_2}(\frac{m_1}{m_1 + m_2})cos(\phi_1 - \phi_2)
\end{equation}

\begin{equation}
    \alpha_2(\phi_2, \phi_2) = \frac{l_1}{l_2}cos(\phi_1 - \phi_2) \label{eq:alpha_equations}
\end{equation}

\begin{equation}
    \funk_1(\phi_1, \phi_2, \dot{\phi}_1, \dot{\phi}_2) = -\frac{l_1}{l_2}(\frac{m_2}{m_1 + m_2})\dot{\phi_2}^2sin(\phi_1 - \phi_2) - \frac{g}{l_1}sin(\phi_1) \label{eq:g1_equation}
\end{equation}

\begin{equation}
    \funk_2(\phi_1, \phi_2, \dot{\phi_1}, \dot{\phi_2}) = \frac{l_1}{l_2}\dot{\phi_1}^2 sin(\phi_1 - \phi_2) - \frac{g}{l_2}sin(\phi_2) \label{eq:g2_equation}
\end{equation}
Where \(\phi_1\) and \(\phi_2\) are the angular positions of the two links of the pendulum, \(m_1\) and \(m_2\) denote the masses at the end of the two links, \(l_1\) and \(l_2\) denote the lengths of the links, and \(g\) is the gravity of Earth.
For a full derivation of these dynamics, we refer the reader to \cite{DiegoAssencio}.

\paragraph{Details of the Symmetry Constraints on \(\funk_1\) and \(\funk_2\) Used as A-Priori Knowledge (K2).}
The four equality constraints included as a-priori knowledge are given by equations \eqref{eq:eq_constr_1}-\eqref{eq:eq_constr_4}.

\begin{align}
\funk_1(\state_{1:2}, \state_{3:4}) = -\funk_1(-\state_{1:2}, \state_{3:4}), \label{eq:eq_constr_1} \\
\funk_2(\state_{1:2}, \state_{3:4}) = -\funk_2(-\state_{1:2}, \state_{3:4}), \label{eq:eq_constr_2}\\
\funk_1(\state_{12}, \state_{3:4}) = \funk_1(\state_{1:2}, -\state_{3:4}), \label{eq:eq_constr_3}\\
\funk_2(\state_{1:2}, \state_{3:4}) = \funk_2(\state_{1:2}, -\state_{3:4}). \label{eq:eq_constr_4}
\end{align}

\paragraph{Generating the Training and Testing Datasets.}
In order to generate training data, the above dynamics were integrated using the Dormand-Prince Runge-Kutta ODE integrator \cite{dormand1980family} implemented within Jax \cite{jax2018github}.
The training dataset consists of a single trajectory of 300 datapoints, with a timestep size of \(\Delta t = 0.01 [s]\).
The testing dataset consists of $10$ trajectories of each with 300 datapoints each and with the same timestep size. 
The initial point of each trajectory used for training is sampled from the intervals \(\phi_1^{init} \in [-0.5, 0]\), \(\phi_2^{init} \in [-0.5,0]\), \(\dot{\phi}_1^{init} \in [-0.3,0.3]\), and \(\dot{\phi}_2^{init} \in [-0.3,0.3]\).
The initial point of each trajectory in the testing dataset is uniformly sampled from the intervals \(\phi_1^{init} \in [-0.5, 0.5]\), \(\phi_2^{init} \in [-0.5,0.5]\), \(\dot{\phi}_1^{init} \in [-0.6,0.6]\), and \(\dot{\phi}_2^{init} \in [-0.6, 0.6]\).
This slight difference in the testing and training datasets helps demonstrate that the incorporation of a-priori knowledge into the model yields better generalization results.

\begin{figure}
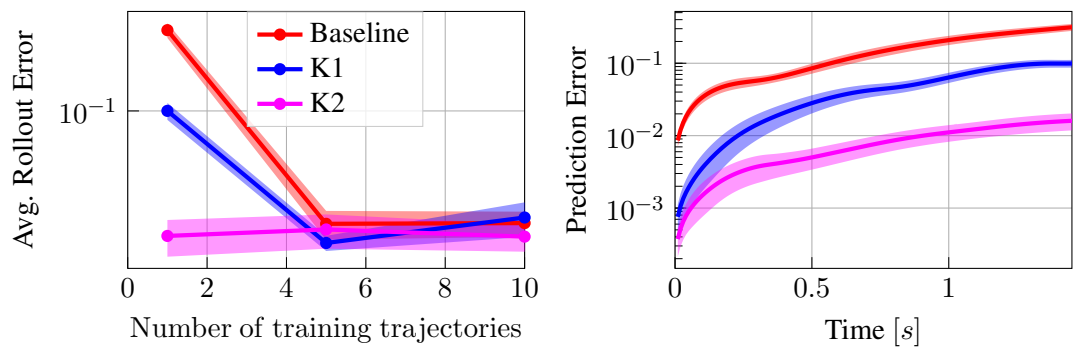

    \centering
    % This file was created with tikzplotlib v0.9.12.
\begin{tikzpicture}

\begin{groupplot}[group style={group name = plots, group size=2 by 1, vertical sep=0pt, horizontal sep=2cm}]

\nextgroupplot[
    width=0.45\textwidth, 
    height=5cm,
    legend cell align={left},
    legend style={
                fill opacity=0.0, 
                anchor=north west,
                at={(0.3, 1.0)},
                draw opacity=1, 
                text opacity=1, 
                draw=white!80!black
                },
    legend cell align={left},
    legend style={fill opacity=0.8, draw opacity=1, text opacity=1, draw=white!80!black},
    log basis y={10},
    tick align=inside,
    tick pos=left,
    x grid style={white!69.0196078431373!black},
    xlabel={\(\displaystyle \mathrm{Number \ of \ training \ trajectories }\)},
    xmajorgrids,
    xmin=0.0, xmax=10,
    xtick style={color=black},
    y grid style={white!69.0196078431373!black},
    ylabel={Avg. Rollout Error},
    ymajorgrids,
    ymin=0.0101485065654793, ymax=0.424396541788506,
    ymode=log,
    ytick style={color=black}
]
\input{supp_mat/double_pend/error_vs_trajectory_count_data}

\nextgroupplot[
    width=0.45\textwidth, 
    height=5cm,
    legend cell align={left},
    legend style={
      fill opacity=0.8,
      draw opacity=1,
      text opacity=1,
      at={(0.03,0.97)},
      anchor=north west,
      draw=white!80!black
    },
    log basis y={10},
    tick align=inside,
    tick pos=left,
    x grid style={white!69.0196078431373!black},
    xlabel={Time \([s]\)},
    xmajorgrids,
    xmin=-0.0, xmax=1.45,
    xtick style={color=black},
    y grid style={white!69.0196078431373!black},
    ylabel={Prediction Error},
    ymajorgrids,
    ymin=0.000146882736422259, ymax=0.519234817976574,
    ymode=log,
    ytick style={color=black}
]
\input{supp_mat/double_pend/error_vs_time_data}

\end{groupplot}

\end{tikzpicture}
    \caption{
        Error comparison of the dynamics models of the double pendulum.
        The \textit{Avg. Rollout Error} plot shows the average error (calculated over a number of short rollouts) as a function of the number of trajectories included in the training dataset.
        The \textit{Prediction Error} plot shows the geometric mean of the model prediction error (calculated over several short rollouts) as a function of time.
        All plots are generated using models whose training has converged.
        }
    \label{fig:supp_double_pend_err_plot}
\end{figure}

\begin{figure}
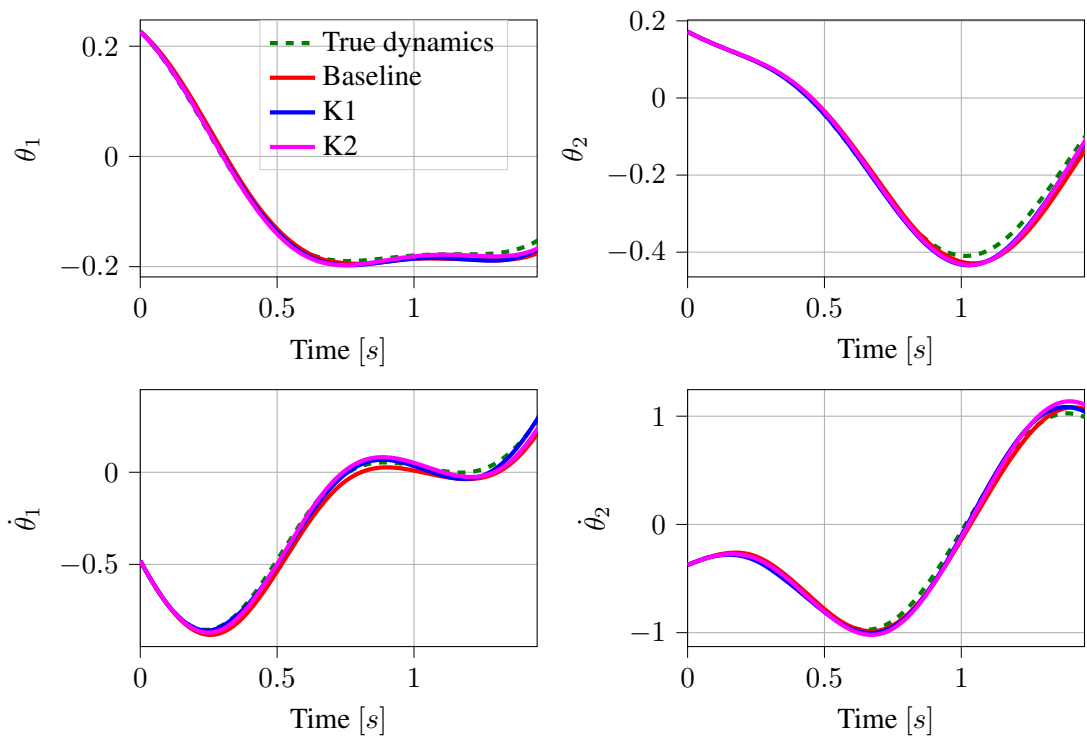

    \centering
    % This file was created with tikzplotlib v0.9.12.
\begin{tikzpicture}

\definecolor{color0}{rgb}{1,0,1}

\begin{groupplot}[group style={group name = plots, group size=2 by 2, vertical sep=1.5cm, horizontal sep=2.0cm}]

%%%%%%%%%%%%%% testing loss vs training steps plot
\nextgroupplot[
width=0.45\textwidth, 
height=5cm,
legend cell align={left},
legend style={
            fill opacity=0.0, 
            anchor=north west,
            at={(0.3, 1.0)},
            draw opacity=1, 
            text opacity=1, 
            draw=white!80!black
            },
tick align=outside,
tick pos=left,
x grid style={white!69.0196078431373!black},
xlabel={Time \([s]\)},
xmajorgrids,
xmin=-0.0, xmax=1.45,
xtick style={color=black},
y grid style={white!69.0196078431373!black},
ylabel={\(\theta_1\)},
ymajorgrids,
ymin=-0.218618423491716, ymax=0.24760492220521,
ytick style={color=black}
]
\input{supp_mat/double_pend/phi_1_dynamics_data}

%%%%%%%%%%%%%%% rollout err vs num training trajectories plot
\nextgroupplot[
width=0.45\textwidth, 
height=5cm,
legend cell align={left},
legend style={fill opacity=0.8, draw opacity=1, text opacity=1, draw=white!80!black},
tick align=outside,
tick pos=left,
x grid style={white!69.0196078431373!black},
xlabel={Time \([s]\)},
xmajorgrids,
xmin=-0.0, xmax=1.45,
xtick style={color=black},
y grid style={white!69.0196078431373!black},
ylabel={\(\theta_2\)},
ymajorgrids,
ymin=-0.464159694314003, ymax=0.202306953072548,
ytick style={color=black}
]
\input{supp_mat/double_pend/phi_2_dynamics_data}

%%%%%%%%%% Prediction error plot
\nextgroupplot[
width=0.45\textwidth, 
height=5cm,
legend cell align={left},
legend style={
  fill opacity=0.8,
  draw opacity=1,
  text opacity=1,
  at={(0.03,0.97)},
  anchor=north west,
  draw=white!80!black
},
tick align=outside,
tick pos=left,
x grid style={white!69.0196078431373!black},
xlabel={Time \([s]\)},
xmajorgrids,
xmin=-0.0, xmax=1.45,
xtick style={color=black},
y grid style={white!69.0196078431373!black},
ylabel={\(\dot{\theta}_1\)},
ymajorgrids,
ymin=-0.946083036065102, ymax=0.448412683606148,
ytick style={color=black}
]
\input{supp_mat/double_pend/phi_1_dot_dynamics_data}

%%%%%%%% CONSTRAINT LOSS PLOT
\nextgroupplot[
width=0.45\textwidth, 
height=5cm,
legend cell align={left},
legend style={
  fill opacity=0.8,
  draw opacity=1,
  text opacity=1,
  at={(0.03,0.97)},
  anchor=north west,
  draw=white!80!black
},
tick align=outside,
tick pos=left,
x grid style={white!69.0196078431373!black},
xlabel={Time \([s]\)},
xmajorgrids,
xmin=-0.0, xmax=1.45,
xtick style={color=black},
y grid style={white!69.0196078431373!black},
ylabel={\(\dot{\theta}_2\)},
ymajorgrids,
ymin=-1.1285467505455, ymax=1.24518374204636,
ytick style={color=black}
]
\input{supp_mat/double_pend/phi_2_dot_dynamics_data}
\end{groupplot}

\end{tikzpicture}
    \caption{Predicted dynamics of the double pendulum system.}
    \label{fig:supp_double_pend_state_plot}
\end{figure}

\paragraph{Additional Experimental Results.}
Figure \ref{fig:supp_double_pend_err_plot} gives further comparisons of the errors of the learned double pendulum dynamics models, while Figure \ref{fig:supp_double_pend_state_plot} illustrates the learned dynamics from a given initial point.

\newpage
\section{Supplementary Robotics Experiments Details and Results}
\label{sec:supp_mat_robotics_details}

\paragraph{Complete Brax Dynamics.} We recall that the dynamics of muti-body rigid systems are in general given by 
\begin{equation}
    \Ddot{q} = \massMat(q)^{-1}[\corForce(q,\dot{q}) + \actuationForce(q, \dot{q}, \controlInput) + \jacobian(q)\mathcal{\contactForce}(q, \dot{q}, \controlInput)],
\end{equation}
where $q$ represents the position or angle of the bodies, and $\dot{q}$ represents the velocity or rotational velocities of the bodies.

However, Brax's implementation of the dynamics is slightly different from those given above. In fact, angular values $q$ are instead represented by quaternions. Specifically, the system' state is given by $q = [\mathrm{pos}, \mathrm{quat}]$ and $\dot{q} = [v, \omega ]$, where $\mathrm{pos}$ is the position, $\mathrm{quat}$ is the quaternion representation of the angular positions, $v$ is the velocity, and $\omega$ is the rotation velocity. Thus, the actual full dynamics of the system are given by:
\begin{align}
    \dot{\mathrm{pos}} = v; \quad \dot{\mathrm{quat}} = \frac{1}{2} [0; \omega] \cdot \mathrm{quat}; \quad \Ddot{q} = \massMat(q)^{-1}[\corForce(q,\dot{q}) + \actuationForce(q, \dot{q}, \controlInput) + \jacobian(q)\mathcal{\contactForce}(q, \dot{q}, \controlInput)],
\end{align} 
where $\cdot$ denotes the product between two quaternion values. 

In this paper, (K1) assumes the a-priori knowledge that $\dot{\mathrm{pos}} = v; \: \dot{\mathrm{quat}} = \frac{1}{2} \omega \cdot \mathrm{quat}$ is given.

\paragraph{Environment Details.}

As each environment in Brax is essentially composed of multiple bodies, some bodies in the environments might be inactive. That is, their states do not evolve in time. For example, the ground in each environment is considered to be inactive. In these settings, we reduce the number of states present in $q$ and $\dot{q}$ by eliminating these variables and learning the dynamics of the active states only. 

\begin{table}[!hbt]
\centering
        \begin{tabular}{lllllll}
            \hline
            
           & \multicolumn{1}{c}{Ant} & \multicolumn{1}{c}{Fetch} &  \multicolumn{1}{c}{Humanoid} & \multicolumn{1}{c}{Reacher} & \multicolumn{1}{c}{UR5E}  \\
           
           \hline
            
            \multicolumn{1}{l}{Number of control inputs} & \multicolumn{1}{l}{8} & \multicolumn{1}{l}{10} &  \multicolumn{1}{l}{17} & \multicolumn{1}{l}{2} & \multicolumn{1}{l}{6}\\
        
            \hline
            
            \multicolumn{1}{l}{Number of active states} & \multicolumn{1}{l}{117} & \multicolumn{1}{l}{143} &  \multicolumn{1}{l}{143} & \multicolumn{1}{l}{12} & \multicolumn{1}{l}{78}\\
            
            \hline
            
            \multicolumn{1}{l}{Total number of states} & \multicolumn{1}{l}{130} & \multicolumn{1}{l}{169} &  \multicolumn{1}{l}{156} & \multicolumn{1}{l}{52} & \multicolumn{1}{l}{104}\\
            
            \hline
            
        \end{tabular}
    \caption{The number of states and control variables in each brax environment.}
    \label{tab:dimension-brax}
\end{table}

\paragraph{Additional Experimental Results.}
We provide the experimental results for all of the robotics environments below.
We note that the \textit{Ant}, \textit{Fetch}, and \textit{Humanoid} experiments feature contact forces, while the \textit{Reacher} and \textit{UR5E Arm} experiments do not.
We accordingly only include the results of the \((K3)\) and \((K4)\) experiments (which incorporate a-priori knowledge pertaining to the contact forces) for the former three environments.

\begin{figure*}[b]
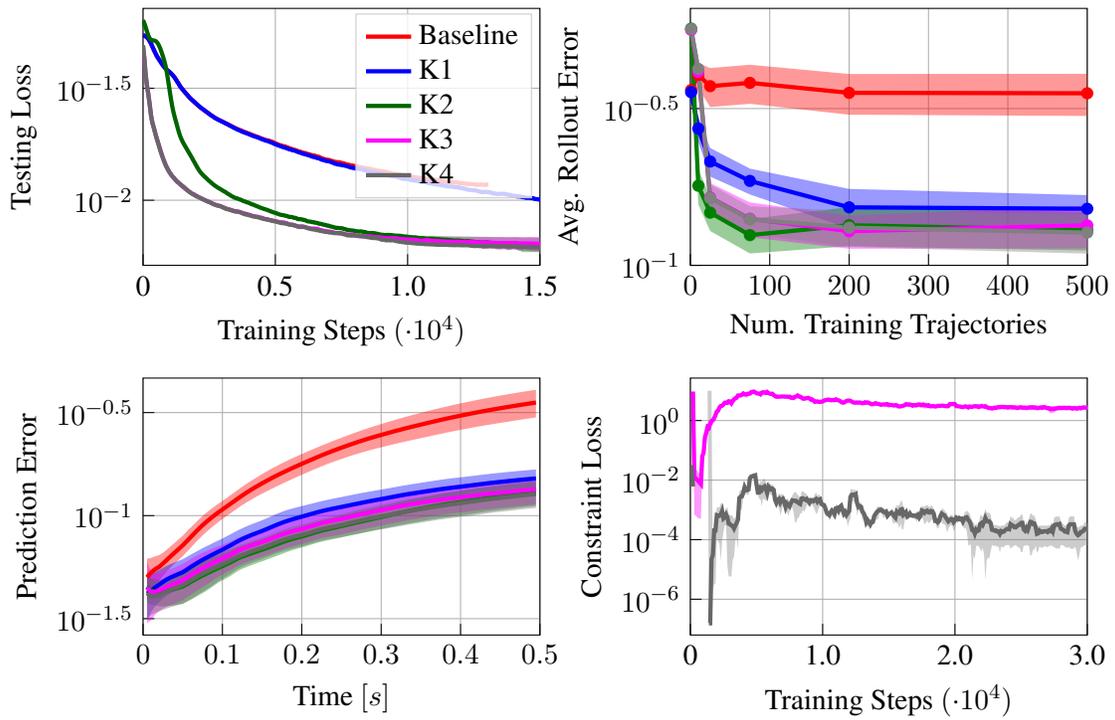

    \centering
    % This file was created with tikzplotlib v0.9.12.
\begin{tikzpicture}

\definecolor{color0}{rgb}{1,0,1}

\begin{groupplot}[group style={group name = plots, group size=2 by 2, vertical sep=1.5cm, horizontal sep=2.0cm}]

%%%%%%%%%%%%%% testing loss vs training steps plot
\nextgroupplot[
width=0.45\textwidth, 
height=5cm,
legend cell align={left},
legend style={fill opacity=0.8, draw opacity=1, text opacity=1, draw=white!80!black},
log basis y={10},
% scaled x ticks=manual:{}{\pgfmathparse{#1}},
tick align=inside,
tick pos=left,
x grid style={white!69.0196078431373!black},
xlabel={Training Steps \((\cdot 10^4)\)},
xmajorgrids,
xmin=-0, xmax=30000 ,
xtick style={color=black},
scaled x ticks=false,
xticklabels={-10, 0, 0.5, 1.0, 1.5, 2.0, 2.5, 3.0},
y grid style={white!69.0196078431373!black},
ylabel={Testing Loss},
ymajorgrids,
ymin=0.00512182982436264, ymax=0.0718282045075842,
ymode=log,
ytick style={color=black},
]
\input{supp_mat/ant_plots/train_test_loss_data}

%%%%%%%%%%%%%%% rollout err vs num training trajectories plot
\nextgroupplot[
width=0.45\textwidth, 
height=5cm,
legend cell align={left},
legend style={fill opacity=0.8, draw opacity=1, text opacity=1, draw=white!80!black},
log basis y={10},
tick align=inside,
tick pos=left,
x grid style={white!69.0196078431373!black},
xlabel={Num. Training Trajectories},
xmajorgrids,
xmin=-0, xmax=500,
xtick style={color=black},
y grid style={white!69.0196078431373!black},
ylabel={Avg. Rollout Error},
ymajorgrids,
ymin=0.100004961961089, ymax=0.659923392230301,
ymode=log,
ytick style={color=black}
]
\input{supp_mat/ant_plots/traj_err_data}

%%%%%%%%%% Prediction error plot
\nextgroupplot[
width=0.45\textwidth, 
height=5cm,
legend cell align={left},
legend style={
  fill opacity=0.8,
  draw opacity=1,
  text opacity=1,
  at={(0.03,0.97)},
  anchor=north west,
  draw=white!80!black
},
log basis y={10},
tick align=inside,
tick pos=left,
x grid style={white!69.0196078431373!black},
xlabel={Time \([s]\)},
xmajorgrids,
xmin=-0.0, xmax=0.5,
xtick style={color=black},
y grid style={white!69.0196078431373!black},
ylabel={Prediction Error},
ymajorgrids,
ymin=0.0263478572825481, ymax=0.465021971504571,
ymode=log,
ytick style={color=black}
]
\input{supp_mat/ant_plots/relerr_data}

%%%%%%%% CONSTRAINT LOSS PLOT
\nextgroupplot[
width=0.45\textwidth, 
height=5cm,
legend cell align={left},
legend style={fill opacity=0.8, draw opacity=1, text opacity=1, draw=white!80!black},
log basis y={10},
tick align=inside,
tick pos=left,
unbounded coords=jump,
x grid style={white!69.0196078431373!black},
xlabel={Training Steps \((\cdot 10^{4})\)},
xmin=0, xmax=30000,
xtick style={color=black},
xtick={-10000,0,10000,20000,30000,40000,50000},
xticklabels={\ensuremath{-}1.0,0,1.0,2.0,3.0,4.0,5.0},
scaled x ticks=false,
y grid style={white!69.0196078431373!black},
ylabel={Constraint Loss},
ymin=6.28126474329451e-08, ymax=26.7046814964445,
xmajorgrids,
ymajorgrids,
ymode=log,
ytick style={color=black},
ytick={1e-10,1e-08,1e-06,0.0001,0.01,1,100,10000},
yticklabels={
  \(\displaystyle {10^{-10}}\),
  \(\displaystyle {10^{-8}}\),
  \(\displaystyle {10^{-6}}\),
  \(\displaystyle {10^{-4}}\),
  \(\displaystyle {10^{-2}}\),
  \(\displaystyle {10^{0}}\),
  \(\displaystyle {10^{2}}\),
  \(\displaystyle {10^{4}}\)
}
]
\input{supp_mat/ant_plots/constraint_loss_data}
\end{groupplot}

\end{tikzpicture}
    \caption{Additional results for the \textit{Ant} experiments.}
    \label{fig:supp_ant_fig}
\end{figure*}

\begin{figure*}[b]
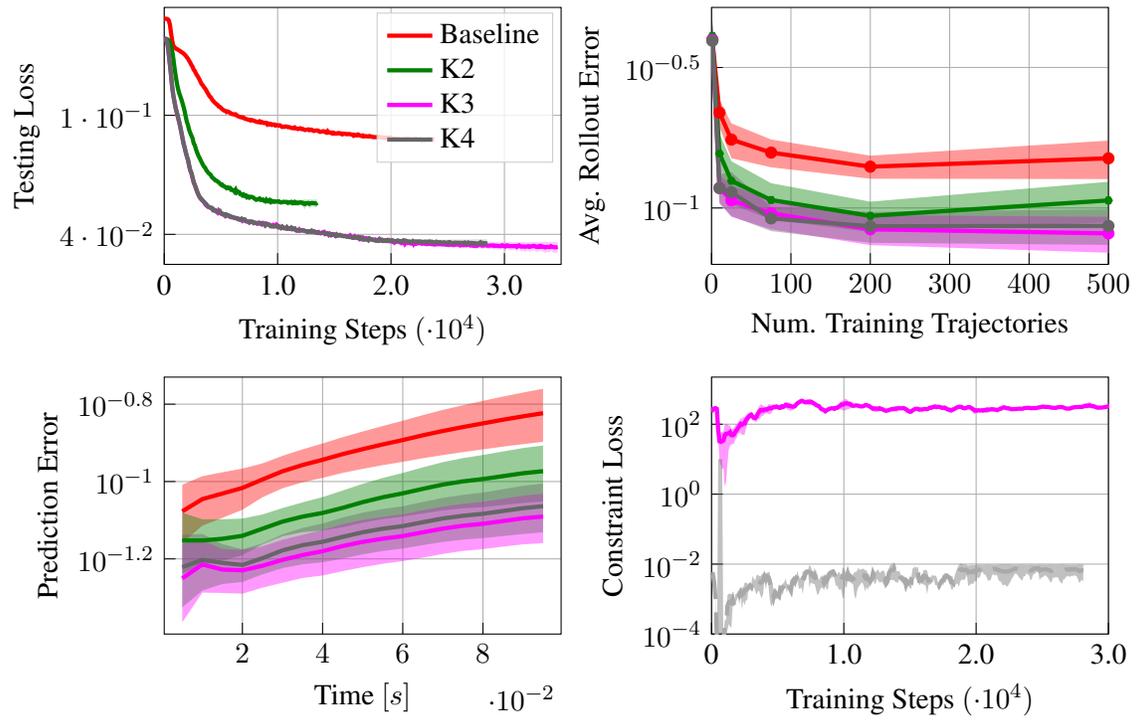

    \centering
    % This file was created with tikzplotlib v0.9.12.
\begin{tikzpicture}

\definecolor{color0}{rgb}{1,0,1}

\begin{groupplot}[group style={group name = plots, group size=2 by 2, vertical sep=1.5cm, horizontal sep=2.0cm}]

%%%%%%%%%%%%%% testing loss vs training steps plot
\nextgroupplot[
width=0.45\textwidth, 
height=5cm,
legend cell align={left},
legend style={fill opacity=0.8, draw opacity=1, text opacity=1, draw=white!80!black},
log basis y={10},
% scaled x ticks=manual:{}{\pgfmathparse{#1}},
tick align=inside,
tick pos=left,
x grid style={white!69.0196078431373!black},
xlabel={Training Steps \((\cdot 10^4)\)},
xmajorgrids,
xmin=0, xmax=35000,
xtick={0, 10000, 20000, 30000},
xticklabels={0, 1.0, 2.0, 3.0},
xtick style={color=black},
scaled x ticks=false,
y grid style={white!69.0196078431373!black},
ylabel={Testing Loss},
ymajorgrids,
ymin=0.0317490094894555, ymax=0.229388645238984,
ytick={0.04, 0.1},
yticklabels={\(4 \cdot 10^{-2}\), \(1 \cdot 10^{-1}\)},
ymode=log,
ytick style={color=black},
]
\input{supp_mat/fetch_plots/train_test_loss_data}

%%%%%%%%%%%%%%% rollout err vs num training trajectories plot
\nextgroupplot[
width=0.45\textwidth, 
height=5cm,
legend cell align={left},
legend style={fill opacity=0.8, draw opacity=1, text opacity=1, draw=white!80!black},
log basis y={10},
tick align=inside,
tick pos=left,
x grid style={white!69.0196078431373!black},
xlabel={Num. Training Trajectories},
xmajorgrids,
xmin=-0, xmax=500,
xtick style={color=black},
y grid style={white!69.0196078431373!black},
ylabel={Avg. Rollout Error},
ymajorgrids,
ymin=0.06299942014023, ymax=0.517269287287668,
ymode=log,
ytick style={color=black}
]
\input{supp_mat/fetch_plots/traj_err_data}

%%%%%%%%%% Prediction error plot
\nextgroupplot[
width=0.45\textwidth, 
height=5cm,
legend cell align={left},
legend style={
  fill opacity=0.8,
  draw opacity=1,
  text opacity=1,
  at={(0.03,0.97)},
  anchor=north west,
  draw=white!80!black
},
log basis y={10},
tick align=inside,
tick pos=left,
x grid style={white!69.0196078431373!black},
xlabel={Time \([s]\)},
xmajorgrids,
xmin=0.000499999988824129, xmax=0.0994999977760017,
xtick style={color=black},
y grid style={white!69.0196078431373!black},
ylabel={Prediction Error},
ymajorgrids,
ymin=0.0403423932359269, ymax=0.186214482340161,
ymode=log,
ytick style={color=black}
]
\input{supp_mat/fetch_plots/relerr_data}

%%%%%%%% CONSTRAINT LOSS PLOT
\nextgroupplot[
width=0.45\textwidth, 
height=5cm,
legend cell align={left},
legend style={fill opacity=0.8, draw opacity=1, text opacity=1, draw=white!80!black},
log basis y={10},
tick align=inside,
tick pos=left,
unbounded coords=jump,
x grid style={white!69.0196078431373!black},
xlabel={Training Steps \((\cdot 10^{4})\)},
xmin=0, xmax=30000,
xtick style={color=black},
xtick={-10000,0,10000,20000,30000,40000,50000},
xticklabels={\ensuremath{-}1.0,0,1.0,2.0,3.0,4.0,5.0},
scaled x ticks=false,
y grid style={white!69.0196078431373!black},
ylabel={Constraint Loss},
ymin=1e-4, ymax=2225.0296872671,
xmajorgrids,
ymajorgrids,
ymode=log,
ytick style={color=black},
ytick={1e-10,1e-08,1e-06,0.0001,0.01,1,100,10000},
yticklabels={
  \(\displaystyle {10^{-10}}\),
  \(\displaystyle {10^{-8}}\),
  \(\displaystyle {10^{-6}}\),
  \(\displaystyle {10^{-4}}\),
  \(\displaystyle {10^{-2}}\),
  \(\displaystyle {10^{0}}\),
  \(\displaystyle {10^{2}}\),
  \(\displaystyle {10^{4}}\)
}
]
\input{supp_mat/fetch_plots/constraint_loss_data}
\end{groupplot}

\end{tikzpicture}
    \caption{Additional results for the \textit{Fetch} experiments.}
    \label{fig:supp_fetch_fig}
\end{figure*}

\begin{figure*}[b]
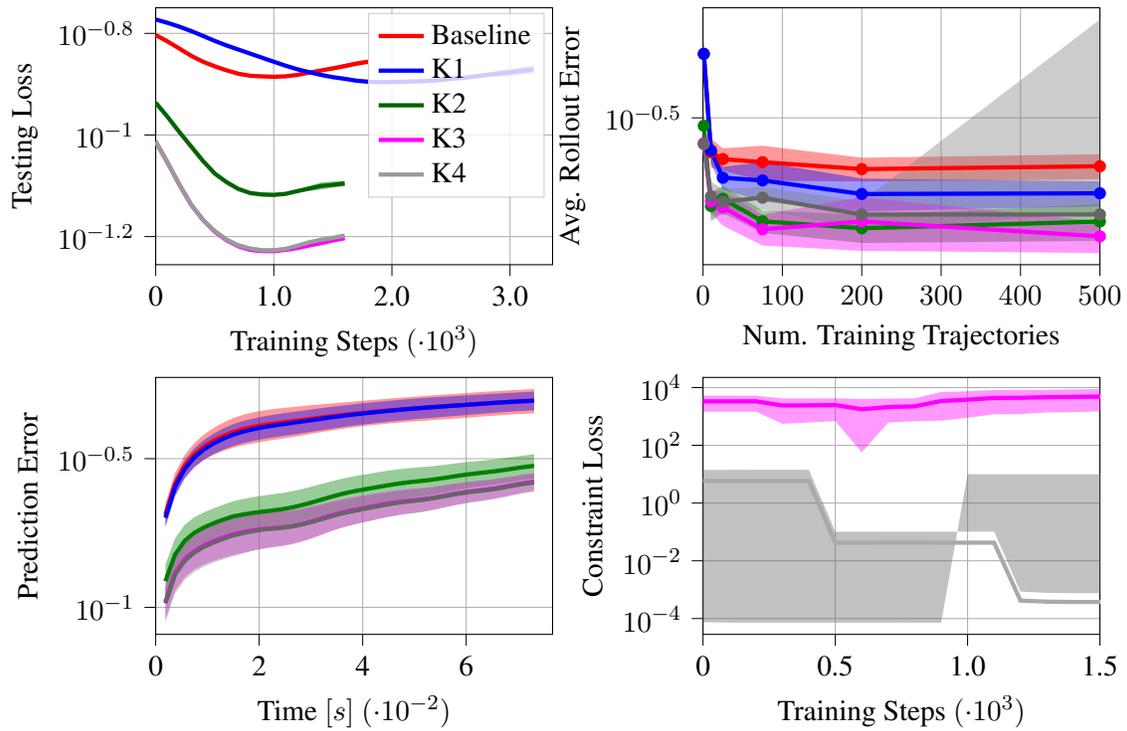

    \centering
    % This file was created with tikzplotlib v0.9.12.
\begin{tikzpicture}

\definecolor{color0}{rgb}{1,0,1}

\begin{groupplot}[group style={group name = plots, group size=2 by 2, vertical sep=1.5cm, horizontal sep=2.0cm}]

%%%%%%%%%%%%%% testing loss vs training steps plot
\nextgroupplot[
width=0.45\textwidth, 
height=5cm,
legend cell align={left},
legend style={fill opacity=0.8, draw opacity=1, text opacity=1, draw=white!80!black},
log basis y={10},
% scaled x ticks=manual:{}{\pgfmathparse{#1}},
tick align=outside,
tick pos=left,
x grid style={white!69.0196078431373!black},
xlabel={Training Steps \((\cdot 10^3)\)},
xmajorgrids,
xmin=-0, xmax=3360,
xtick style={color=black},
scaled x ticks=false,
xtick={0, 1000, 2000, 3000},
xticklabels={0, 1.0, 2.0, 3.0},
y grid style={white!69.0196078431373!black},
ylabel={Testing Loss},
ymajorgrids,
ymin=0.0555508527045486, ymax=0.178010585505123,
ymode=log,
ytick style={color=black},
]
\input{supp_mat/humanoid_plots/train_test_loss_data}

%%%%%%%%%%%%%%% rollout err vs num training trajectories plot
\nextgroupplot[
width=0.45\textwidth, 
height=5cm,
legend cell align={left},
legend style={fill opacity=0.8, draw opacity=1, text opacity=1, draw=white!80!black},
log basis y={10},
tick align=outside,
tick pos=left,
x grid style={white!69.0196078431373!black},
xlabel={Num. Training Trajectories},
xmajorgrids,
xmin=0.0, xmax=500,
xtick style={color=black},
y grid style={white!69.0196078431373!black},
ylabel={Avg. Rollout Error},
ymajorgrids,
ymin=0.112838589141192, ymax=0.684740490837029,
ymode=log,
ytick style={color=black}
]
\input{supp_mat/humanoid_plots/traj_err_data}

%%%%%%%%%% Prediction error plot
\nextgroupplot[
width=0.45\textwidth, 
height=5cm,
legend cell align={left},
legend style={
  fill opacity=0.8,
  draw opacity=1,
  text opacity=1,
  at={(0.03,0.97)},
  anchor=north west,
  draw=white!80!black
},
log basis y={10},
tick align=outside,
tick pos=left,
x grid style={white!69.0196078431373!black},
xlabel={Time \([s]\) \((\cdot 10^{-2})\)},
scaled x ticks=false,
xtick={0, 0.02, 0.04, 0.06},
xticklabels={0, 2, 4, 6},
xmajorgrids,
xmin=-0.0, xmax=0.0766874982859008,
xtick style={color=black},
y grid style={white!69.0196078431373!black},
ylabel={Prediction Error},
ymajorgrids,
ymin=0.0811653602949658, ymax=0.592996093311961,
ymode=log,
ytick style={color=black}
]
\input{supp_mat/humanoid_plots/relerr_data}

%%%%%%%% CONSTRAINT LOSS PLOT
\nextgroupplot[
width=0.45\textwidth, 
height=5cm,
legend cell align={left},
legend style={fill opacity=0.8, draw opacity=1, text opacity=1, draw=white!80!black},
log basis y={10},
tick align=outside,
tick pos=left,
unbounded coords=jump,
x grid style={white!69.0196078431373!black},
xlabel={Training Steps \((\cdot 10^{3})\)},
xmin=0, xmax=1500,
xtick style={color=black},
xtick={0,500, 1000, 1500},
xticklabels={0,0.5, 1.0, 1.5},
scaled x ticks=false,
y grid style={white!69.0196078431373!black},
ylabel={Constraint Loss},
ymin=2.83294982523444e-05, ymax=22479.6659221875,
xmajorgrids,
ymajorgrids,
ymode=log,
ytick style={color=black},
ytick={1e-10,1e-08,1e-06,0.0001,0.01,1,100,10000},
yticklabels={
  \(\displaystyle {10^{-10}}\),
  \(\displaystyle {10^{-8}}\),
  \(\displaystyle {10^{-6}}\),
  \(\displaystyle {10^{-4}}\),
  \(\displaystyle {10^{-2}}\),
  \(\displaystyle {10^{0}}\),
  \(\displaystyle {10^{2}}\),
  \(\displaystyle {10^{4}}\)
}
]
\input{supp_mat/humanoid_plots/constraint_loss_data}
\end{groupplot}

\end{tikzpicture}
    \caption{Additional results for the \textit{Humanoid} experiments.}
    \label{fig:supp_humanoid_fig}
\end{figure*}

\begin{figure*}[b]
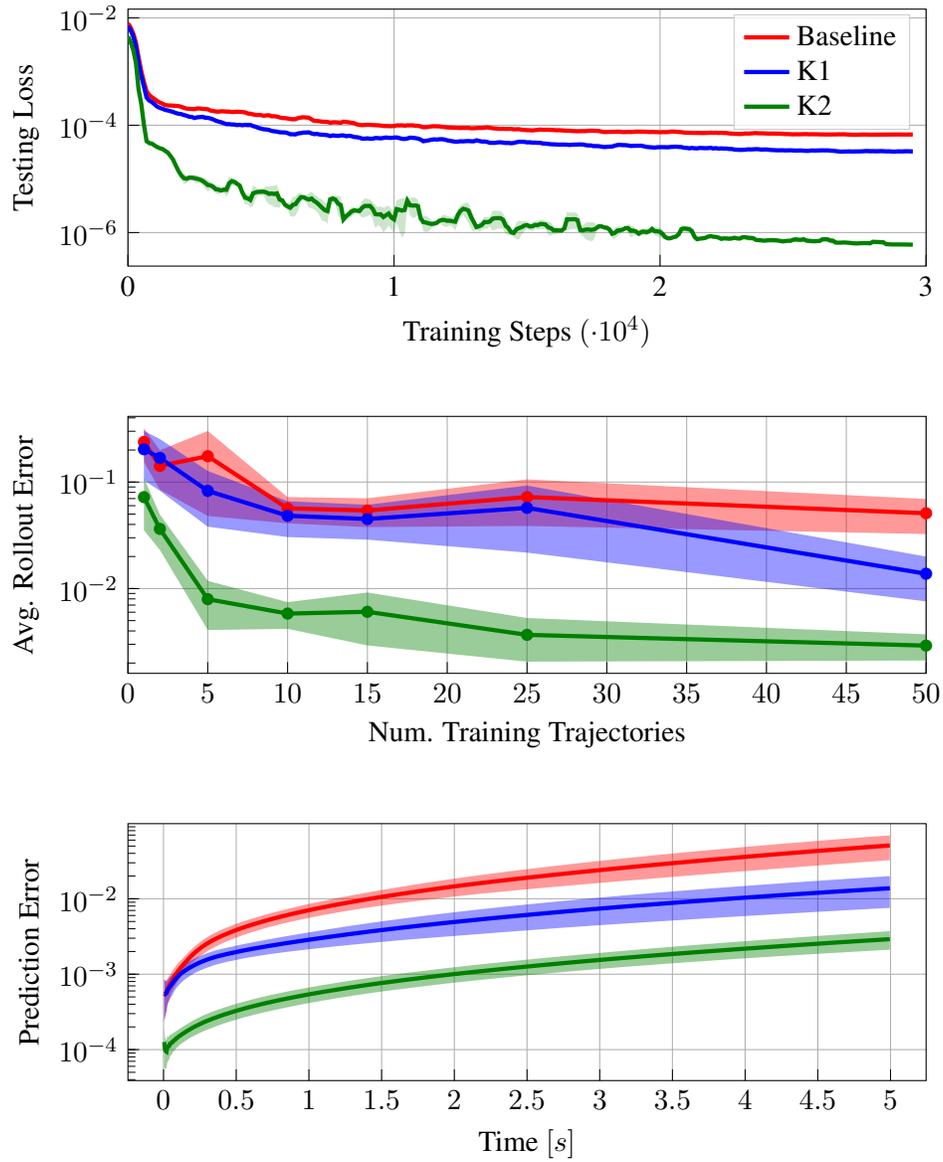

    \centering
    % This file was created with tikzplotlib v0.9.12.
\begin{tikzpicture}

\definecolor{color0}{rgb}{1,0,1}

\begin{groupplot}[group style={group name = plots, group size=1 by 3, vertical sep=2.0cm, horizontal sep=0.0cm}]

%%%%%%%%%%%%%% testing loss vs training steps plot
\nextgroupplot[
width=0.8\textwidth, 
height=5cm,
legend cell align={left},
legend style={fill opacity=0.8, draw opacity=1, text opacity=1, draw=white!80!black},
log basis y={10},
% scaled x ticks=manual:{}{\pgfmathparse{#1}},
tick align=inside,
tick pos=left,
x grid style={white!69.0196078431373!black},
xlabel={Training Steps \((\cdot 10^4)\)},
xmajorgrids,
xmin=-0, xmax=30000,
xtick style={color=black},
scaled x ticks=false,
xtick={0,10000,20000,30000},
xticklabels={0, 1, 2, 3},
y grid style={white!69.0196078431373!black},
ylabel={Testing Loss},
ymajorgrids,
ymin=2.37502224632847e-07, ymax=0.014559881119892,
ymode=log,
ytick style={color=black},
]
\input{supp_mat/reacher_plots/train_test_loss_data}

%%%%%%%%%%%%%%% rollout err vs num training trajectories plot
\nextgroupplot[
width=0.8\textwidth, 
height=5cm,
legend cell align={left},
legend style={fill opacity=0.8, draw opacity=1, text opacity=1, draw=white!80!black},
log basis y={10},
tick align=inside,
tick pos=left,
x grid style={white!69.0196078431373!black},
xlabel={Num. Training Trajectories},
xmajorgrids,
xmin=-0, xmax=50,
xtick style={color=black},
y grid style={white!69.0196078431373!black},
ylabel={Avg. Rollout Error},
ymajorgrids,
ymin=0.00160017906592065, ymax=0.413608681044119,
ymode=log,
ytick style={color=black}
]
\input{supp_mat/reacher_plots/traj_err_data}

%%%%%%%%%% Prediction error plot
\nextgroupplot[
width=0.8\textwidth, 
height=5cm,
legend cell align={left},
legend style={
  fill opacity=0.8,
  draw opacity=1,
  text opacity=1,
  at={(0.03,0.97)},
  anchor=north west,
  draw=white!80!black
},
log basis y={10},
tick align=inside,
tick pos=left,
x grid style={white!69.0196078431373!black},
xlabel={Time \([s]\)},
xmajorgrids,
xmin=-0.244499994534999, xmax=5.24449988277629,
xtick style={color=black},
y grid style={white!69.0196078431373!black},
ylabel={Prediction Error},
ymajorgrids,
ymin=3.88412507472358e-05, ymax=0.0991128189085625,
ymode=log,
ytick style={color=black}
]
\input{supp_mat/reacher_plots/relerr_data}

\end{groupplot}

\end{tikzpicture}
    \caption{Additional results for the \textit{Reacher} experiments.}
    \label{fig:supp_reacher_fig}
\end{figure*}

\begin{figure*}[b]
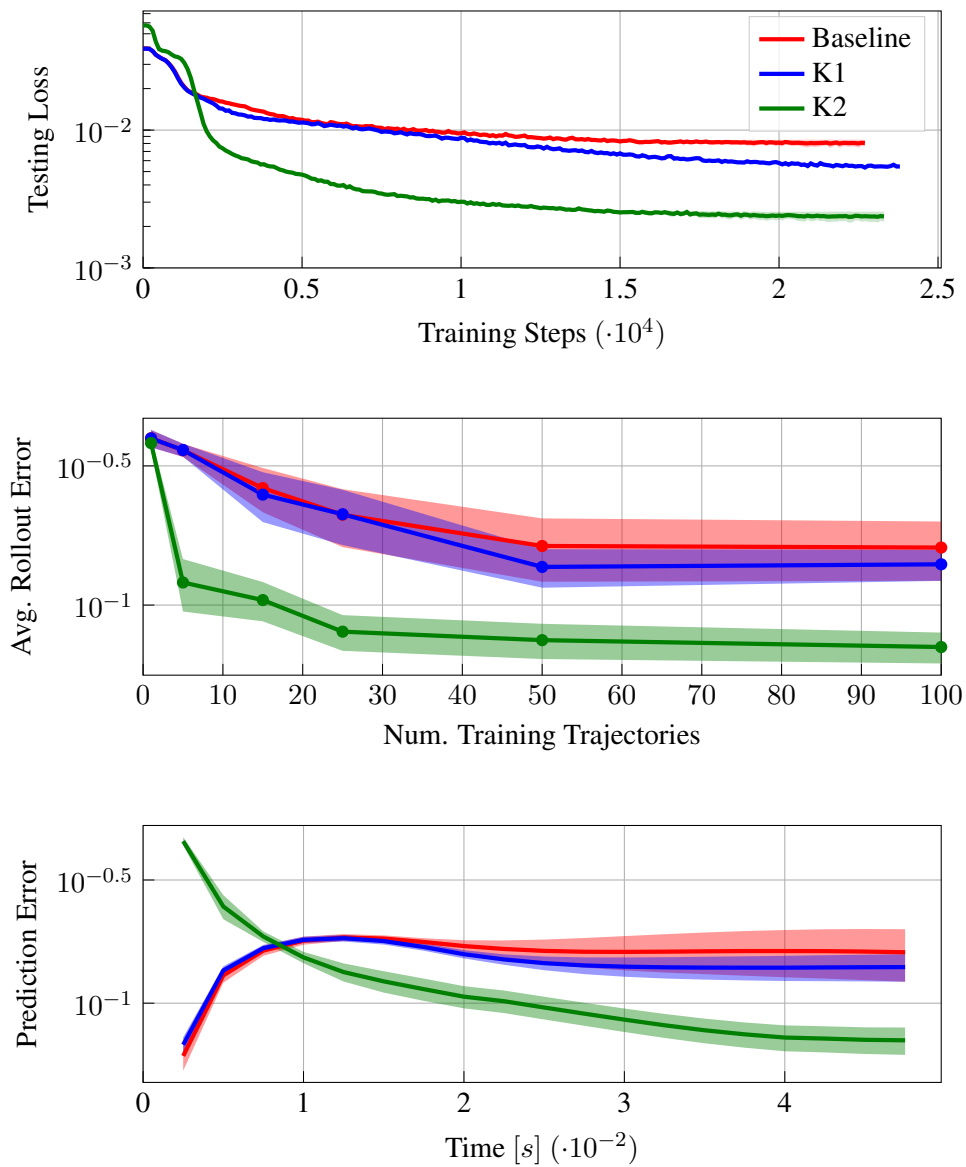

    \centering
    % This file was created with tikzplotlib v0.9.12.
\begin{tikzpicture}

\definecolor{color0}{rgb}{1,0,1}

\begin{groupplot}[group style={group name = plots, group size=1 by 3, vertical sep=2.0cm, horizontal sep=0.0cm}]

%%%%%%%%%%%%%% testing loss vs training steps plot
\nextgroupplot[
width=0.8\textwidth, 
height=5cm,
legend cell align={left},
legend style={fill opacity=0.8, draw opacity=1, text opacity=1, draw=white!80!black},
log basis y={10},
% scaled x ticks=manual:{}{\pgfmathparse{#1}},
tick align=inside,
tick pos=left,
x grid style={white!69.0196078431373!black},
xlabel={Training Steps \((\cdot 10^4)\)},
xmajorgrids,
xmin=-0, xmax=25100,
xtick style={color=black},
scaled x ticks=false,
xtick={0, 5000, 10000, 15000, 20000,25000},
xticklabels={0, 0.5, 1, 1.5, 2, 2.5},
y grid style={white!69.0196078431373!black},
ylabel={Testing Loss},
ymajorgrids,
ymin=0.001, ymax=0.0731729519922072,
ymode=log,
ytick style={color=black},
]
\input{supp_mat/ur5e_plots/train_test_loss_data}

%%%%%%%%%%%%%%% rollout err vs num training trajectories plot
\nextgroupplot[
width=0.8\textwidth, 
height=5cm,
legend cell align={left},
legend style={fill opacity=0.8, draw opacity=1, text opacity=1, draw=white!80!black},
log basis y={10},
tick align=inside,
tick pos=left,
x grid style={white!69.0196078431373!black},
xlabel={Num. Training Trajectories},
xmajorgrids,
xmin=-0, xmax=100,
xtick style={color=black},
y grid style={white!69.0196078431373!black},
ylabel={Avg. Rollout Error},
ymajorgrids,
ymin=0.0560056617060424, ymax=0.469031997291035,
ymode=log,
ytick style={color=black}
]
\input{supp_mat/ur5e_plots/traj_err_data}

%%%%%%%%%% Prediction error plot
\nextgroupplot[
width=0.8\textwidth, 
height=5cm,
legend cell align={left},
legend style={
  fill opacity=0.8,
  draw opacity=1,
  text opacity=1,
  at={(0.03,0.97)},
  anchor=north west,
  draw=white!80!black
},
log basis y={10},
tick align=inside,
tick pos=left,
x grid style={white!69.0196078431373!black},
xlabel={Time \([s]\) \((\cdot 10^{-2})\)},
xmajorgrids,
xmin=0.00, xmax=0.0497499988880008,
scaled x ticks=false,
xtick={0, 0.01, 0.02, 0.03, 0.04},
xticklabels={0, 1, 2, 3, 4},
xtick style={color=black},
y grid style={white!69.0196078431373!black},
ylabel={Prediction Error},
ymajorgrids,
ymin=0.0476849183628596, ymax=0.526339518019372,
ymode=log,
ytick style={color=black}
]
\input{supp_mat/ur5e_plots/relerr_data}

\end{groupplot}

\end{tikzpicture}
    \caption{Additional results for the \textit{UR5E} experiments.}
    \label{fig:supp_ur5e_fig}
\end{figure*}

\end{document}